\documentclass[preprint,5p,times,authoryear,lefttitle]{elsarticle}

\usepackage{amssymb}
\usepackage{amsthm}
\theoremstyle{definition}
\newtheorem{definition}{Definition}

\usepackage{multirow}
\usepackage{subcaption}
\usepackage{booktabs}
\usepackage{amsmath}

\usepackage{algorithm}
\usepackage{algpseudocode}
\algrenewcommand\algorithmicindent{0.9em}
\makeatletter
\let\OldStatex\Statex
\renewcommand{\Statex}[1][3]{%
  \setlength\@tempdima{\algorithmicindent}%
  \OldStatex\hskip\dimexpr#1\@tempdima\relax}
\makeatother

\usepackage{hyperref}
\usepackage[dvipsnames]{xcolor}
\usepackage{arydshln}

\journal{Neural Networks}

\begin{document}

\begin{frontmatter}



\title{MECCH: Metapath Context Convolution-based Heterogeneous Graph Neural Networks}


\author[inst1]{Xinyu Fu}
\author[inst1]{Irwin King}

\affiliation[inst1]{organization={Department of Computer Science and Engineering, The Chinese University of Hong Kong},
            state={Hong Kong},
            country={China}}



\begin{abstract}
Heterogeneous graph neural networks (HGNNs) were proposed for representation learning on structural data with multiple types of nodes and edges.
To deal with the performance degradation issue when HGNNs become deep, researchers combine metapaths into HGNNs to associate nodes closely related in semantics but far apart in the graph.
However, existing metapath-based models suffer from either information loss or high computation costs.
To address these problems, we present a novel \emph{Metapath Context Convolution-based Heterogeneous Graph Neural Network} (MECCH).
{MECCH leverages \emph{metapath contexts}, a new kind of graph structure that facilitates lossless node information aggregation while avoiding any redundancy.}
Specifically, MECCH applies three novel components after feature preprocessing to extract comprehensive information from the input graph efficiently: (1) metapath context construction, (2) metapath context encoder, and (3) convolutional metapath fusion.
Experiments on five real-world heterogeneous graph datasets for node classification and link prediction show that MECCH achieves superior prediction accuracy compared with state-of-the-art baselines with improved computational efficiency.
{The code is available at \url{https://github.com/cynricfu/MECCH}. The formal publication is available at \url{https://doi.org/10.1016/j.neunet.2023.11.030}.}
\end{abstract}




\end{frontmatter}


\section{Introduction}

\begin{figure}[tb]
    \centering
    \includegraphics[width=0.9\columnwidth]{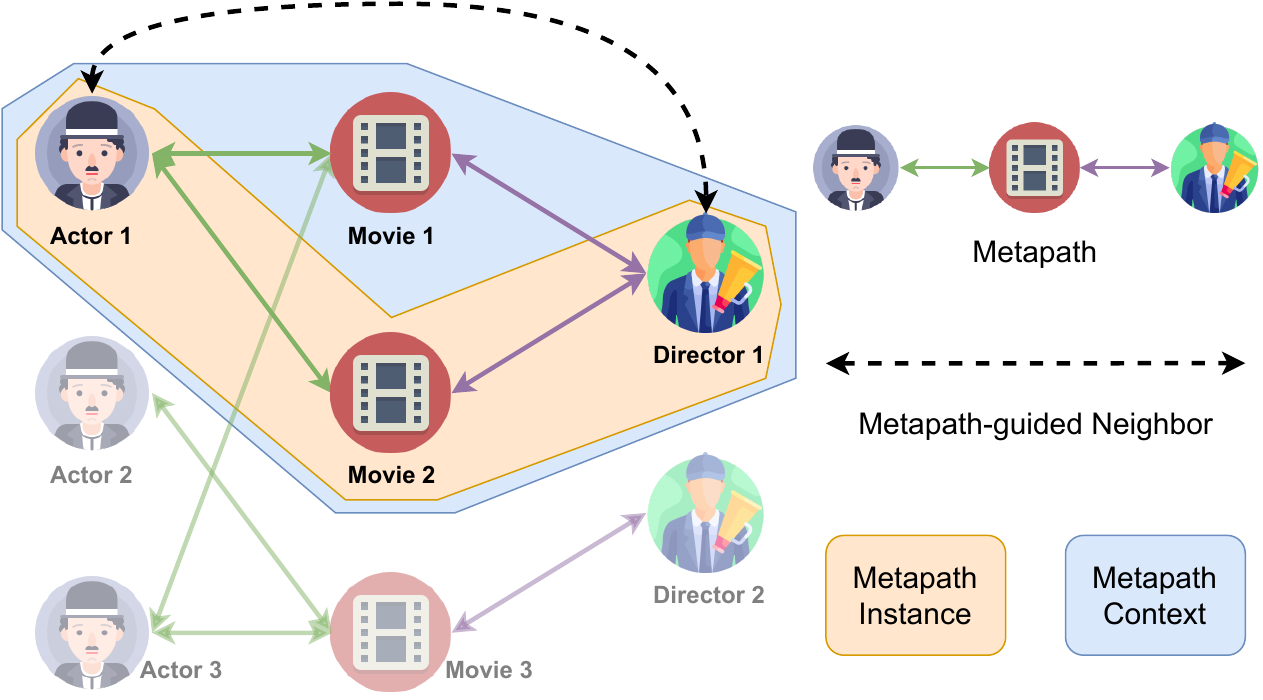}
    \caption{An illustration of the heterogeneous graph and related concepts.}
    \label{fig:teaser}
\end{figure}

Many real-world networks are \emph{heterogeneous graphs}, which contain multiple types of nodes and edges.
As illustrated in Figure~\ref{fig:teaser}, a movie information network may consist of actors, movies, directors, and different types of relationships between them.
The complex and irregular interactions among different types of nodes and edges make it challenging to extract knowledge from heterogeneous graphs efficiently.
Therefore, heterogeneous graph representation learning, which aims to represent nodes using low-dimensional vectors, is a desirable way to automatically process and make inferences on such data.

Over the past decade, heterogeneous graph representation learning has drawn significant attention.
Early attempts usually combine skip-gram model~\citep{DBLP:journals/corr/abs-1301-3781} and metapath-guided random walks~\citep{DBLP:conf/kdd/DongCS17, DBLP:conf/cikm/FuLL17, DBLP:journals/tkde/ShiHZY19}.
With the rapid development of deep learning, graph neural networks (GNNs)~{\citep{DBLP:conf/iclr/KipfW17,DBLP:conf/nips/HamiltonYL17,DBLP:conf/iclr/VelickovicCCRLB18}} are proposed to incorporate node features and benefit from neural network architectures.
Initially, GNNs focused on homogeneous graphs.
But it is straightforward for researchers to generalize GNNs to the heterogeneous scenario, where multiple types of nodes and edges introduce another layer of complexity into the GNN design.

Recent efforts in heterogeneous GNNs (HGNNs) can be divided into two categories: \emph{relation-based HGNNs} and \emph{metapath-based HGNNs}.
Relation-based HGNNs consider message passing parameterized by edge types and aggregate information from direct neighbors~\citep{DBLP:conf/esws/SchlichtkrullKB18,DBLP:conf/kdd/ZhangSHSC19,DBLP:conf/www/HuDWS20,DBLP:conf/kdd/LvDLCFHZJDT21,DBLP:journals/tkde/YangGLZCW23}.
Models of this kind are generally simple and fast. Still, they usually require stacking many GNN layers to leverage information multiple hops away, potentially deteriorating model performance~\citep{DBLP:conf/aaai/LiHW18,DBLP:conf/cikm/ZhouDWLHXF21}.
Another line of research takes advantage of \emph{metapaths}, which are ordered sequences of node and edge types describing composite relationships between nodes (as shown in Figure~\ref{fig:teaser}).
Metapath-based HGNNs consider message passing through metapaths and aggregate information from metapath-guided neighbors.
Through this way, the models can comfortably obtain information multiple hops away with very few layers and capture high-level semantic information embedded in the graph.

However, existing metapath-based HGNNs can hardly achieve a good balance between model performance and computational efficiency.
HAN~\citep{DBLP:conf/www/WangJSWYCY19} aggregates metapath-guided neighbors, but with information loss caused by discarded intermediate nodes along metapaths.
MAGNN~\citep{DBLP:conf/www/0004ZMK20} addresses this issue by encoding metapath instances. But the computational complexity is significantly increased due to redundant computations introduced.
Unlike HAN and MAGNN, which choose metapaths based on human expertise, GTN~\citep{DBLP:conf/nips/YunJKKK19} tries to automatically select informative metapaths using learnable weights. However, GTN also suffers from high computation costs and massive memory usage since it multiplies the entire adjacency matrix of the input graph.

These problems with existing models drive us to design a new metapath-based HGNN that can extract comprehensive knowledge from the input graph while maintaining modest computation time and memory footprint.
{Through analysis, we discover that \emph{metapath contexts}, a new kind of graph structure defined by us, can enable lossless node information aggregation without redundant computations.}
In this paper, we propose a novel \emph{\underline{ME}tapath \underline{C}ontext \underline{C}onvolution-based \underline{H}eterogeneous Graph Neural Network} (MECCH) to learn node representations from heterogeneous graphs.
Specifically, after feature preprocessing, MECCH first constructs metapath contexts for each node to describe the neighboring structures formed by the metapaths.
Next, MECCH efficiently encodes the constructed metapath contexts to integrate information from both metapath-guided neighbors and intermediate nodes along metapaths.
In the end, MECCH adaptively combines node representations from different metapaths using an efficient 1-D convolution kernel.
In summary, our main contributions are as follows:
\begin{enumerate}
    \item We provide a unified framework for metapath-based HGNNs and analyze their limitations, facilitating new HGNN design with in-depth model understanding.
    \item {We introduce metapath contexts, a new kind of graph structure that facilitates lossless node information aggregation while avoiding any redundancy.}
    \item We propose a novel metapath context convolution-based HGNN (MECCH) that can capture comprehensive information without incurring redundant computations.
    \item We conduct extensive experiments on five heterogeneous graph datasets for node classification and link prediction against multiple state-of-the-art HGNNs. MECCH consistently outperforms these baselines while improving computational efficiency.
\end{enumerate}

\paragraph{Paper Structure}
Section~\ref{sec:related_work} briefly introduces the historical development of heterogeneous graph representation learning and especially heterogeneous GNNs.
Section~\ref{sec:preliminary} presents the definitions and graphical illustrations of the important terminologies used in this paper.
Section~\ref{sec:methodology} formulates a general framework for metapath-based HGNNs and presents the novel HGNN model we propose for effective and efficient learning on heterogeneous graphs.
Section~\ref{sec:experiments} describes the experimental settings and evaluation results on the node classification and link prediction tasks.
Section~\ref{sec:conclusion} concludes the paper with possible future directions.

\section{Related Work} \label{sec:related_work}

Heterogeneous graph representation learning aims to transform nodes in a heterogeneous graph into low-dimensional vector representations that can preserve the rich semantic information from the original graph structure and node features.

\subsection{Shallow Models}
Inspired by the homogeneous graph embedding model DeepWalk~\citep{DBLP:conf/kdd/PerozziAS14}, metapath2vec~\citep{DBLP:conf/kdd/DongCS17} generates random walks guided by a pre-defined metapath, which are then fed to a skip-gram model to generate node embeddings.
HIN2vec~\citep{DBLP:conf/cikm/FuLL17} and HERec~\citep{DBLP:journals/tkde/ShiHZY19} also leverage metapaths in different ways.
Some models capture node proximity at the neighborhood level~\citep{DBLP:conf/kdd/ChenYWWNL18} or at the schema level~\citep{DBLP:conf/ijcai/ZhaoWSLY20}.
Some other embedding models focus on knowledge graphs, i.e., heterogeneous graphs with rich schemas~\citep{DBLP:conf/nips/BordesUGWY13,DBLP:journals/corr/YangYHGD14a,DBLP:conf/iclr/SunDNT19}.
{There are also models that conduct heterogeneous graph embedding in non-Euclidean vector spaces~\citep{DBLP:conf/aaai/WangZS19a,DBLP:conf/aaai/WangGHLMV21}.}
These shallow models do not utilize node features, resulting in a heavy loss of important information. They are also trained in an unsupervised fashion. Although their generated embeddings can be fed to downstream models for various tasks, there is a performance gap between them and those end-to-end task-dependent models, i.e., GNNs.


\subsection{Graph Neural Networks}
GNNs have been a recent trend in graph representation learning.
The rationale behind GNNs is that each node can be characterized by its features and local neighborhood.
Following this idea, GCN~\citep{DBLP:conf/iclr/KipfW17}, GAT~\citep{DBLP:conf/iclr/VelickovicCCRLB18}, and many other powerful GNN architectures~\citep{DBLP:conf/nips/HamiltonYL17,DBLP:conf/uai/ZhangSXMKY18,DBLP:conf/ijcai/ZhangSZK19,DBLP:conf/icml/YangLWZGWZHW21} have been proposed.
GNNs can leverage node features and benefit from various training paradigms of deep learning, achieving state-of-the-art results in many homogeneous graph datasets~\citep{DBLP:journals/corr/abs-2206-08181,DBLP:conf/kdd/00010PK23,DBLP:journals/tkde/YangZXK23,DBLP:conf/icml/00010Y0K23} with various applications, such as recommendation~\citep{chen2023star,DBLP:conf/sigir/0001Z0S0K23}, biochemistry~\citep{DBLP:conf/sdm/MengLZ0K23,DBLP:conf/ijcai/MengZ0K23,DBLP:conf/ijcai/MengZ0K23a}, and natural language~\citep{DBLP:conf/aaai/SongK22,DBLP:conf/aaai/MaSHLZK23}. It is natural for researchers to generalize GNNs to the heterogeneous setting.
We summarize recent HGNN designs into relation-based models and metapath-based models.

\begin{figure*}[t]
    \centering
    \begin{subfigure}[b]{0.2\textwidth}
        \centering
        \includegraphics[width=\textwidth]{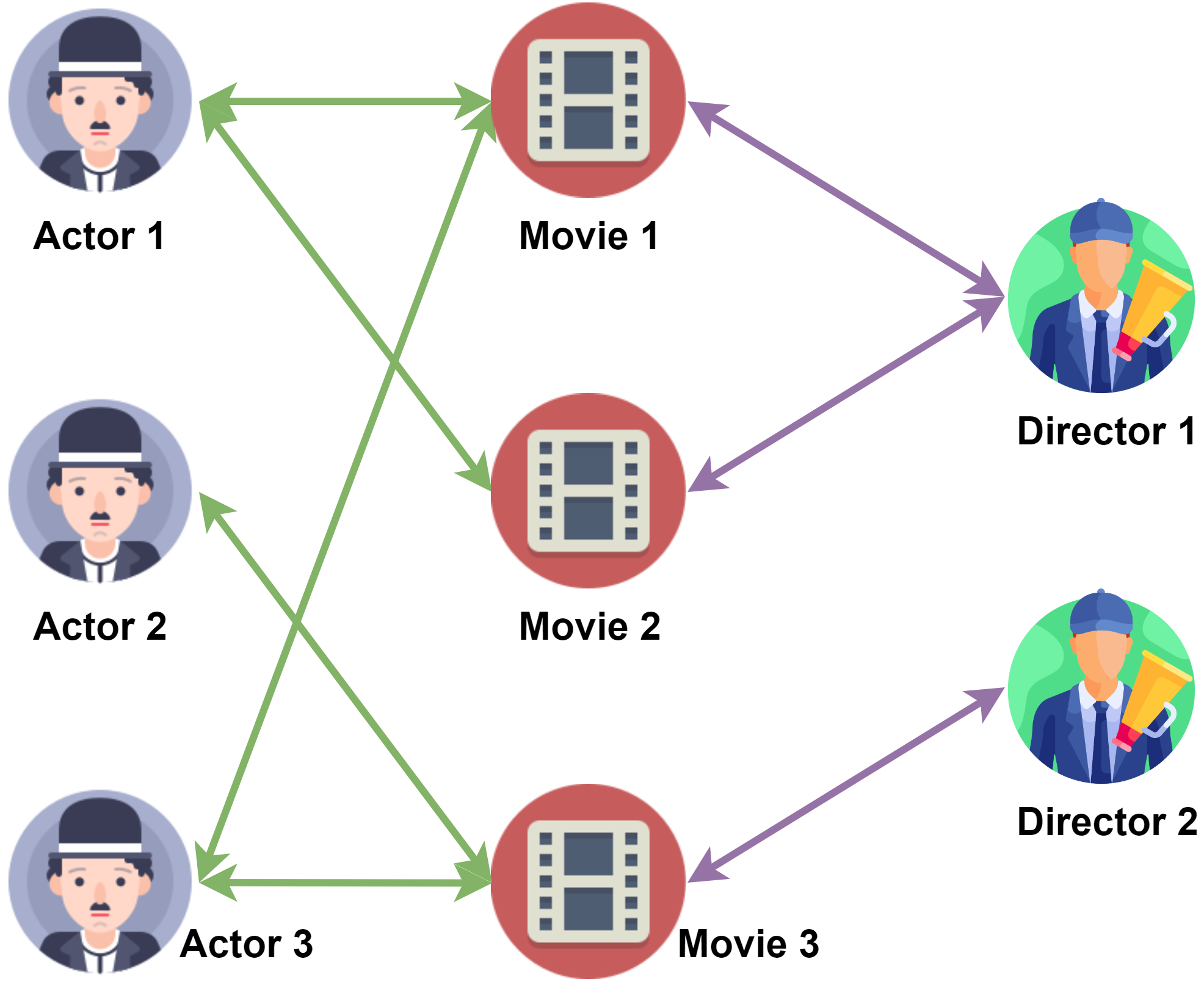}
        \caption{Heterogeneous Graph}
    \end{subfigure}
    \hfill
    \begin{tabular}[b]{@{}c@{}}
        \begin{subfigure}[b]{0.24\textwidth}
            \centering
            \includegraphics[width=\textwidth]{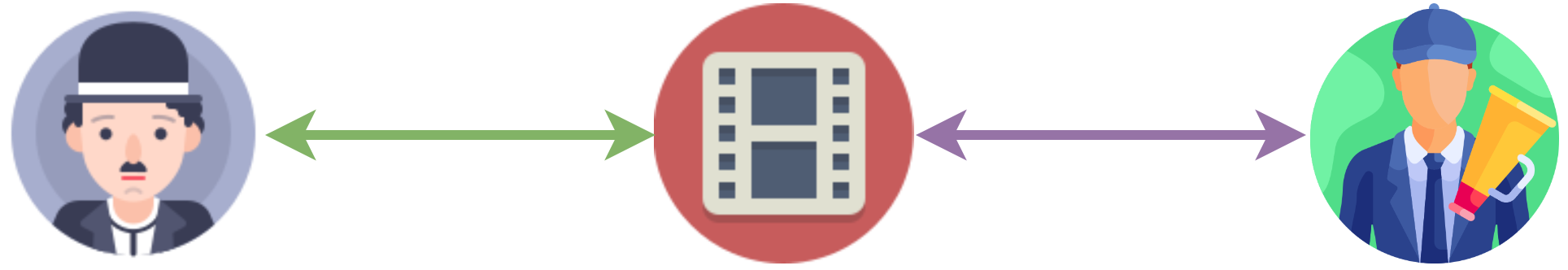}
            \caption{Metapath}
        \end{subfigure} \\
        \begin{subfigure}[b]{0.24\textwidth}
            \centering
            \includegraphics[width=\textwidth]{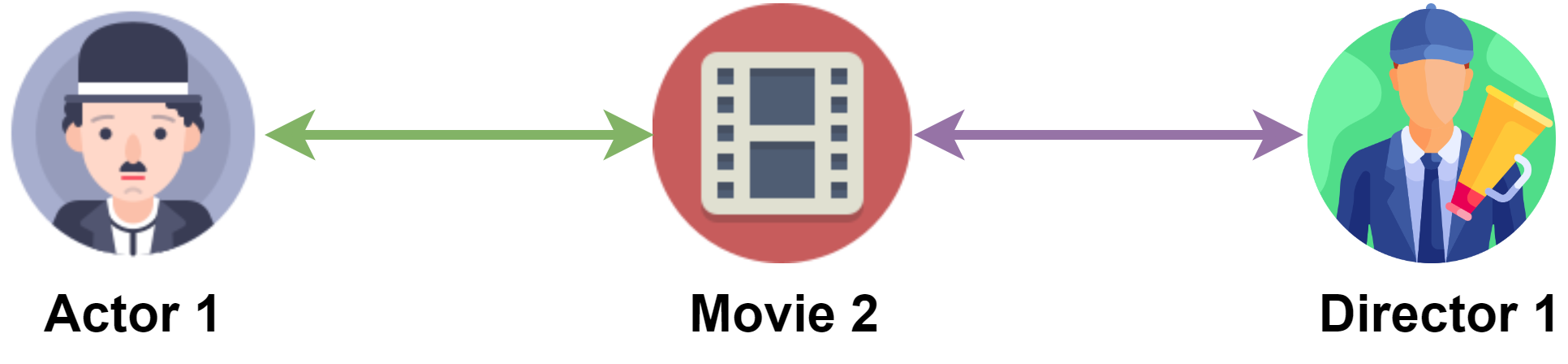}
            \caption{Metapath Instance}
        \end{subfigure}
    \end{tabular}
    \hfill
    \begin{subfigure}[b]{0.24\textwidth}
        \centering
        \includegraphics[width=\textwidth]{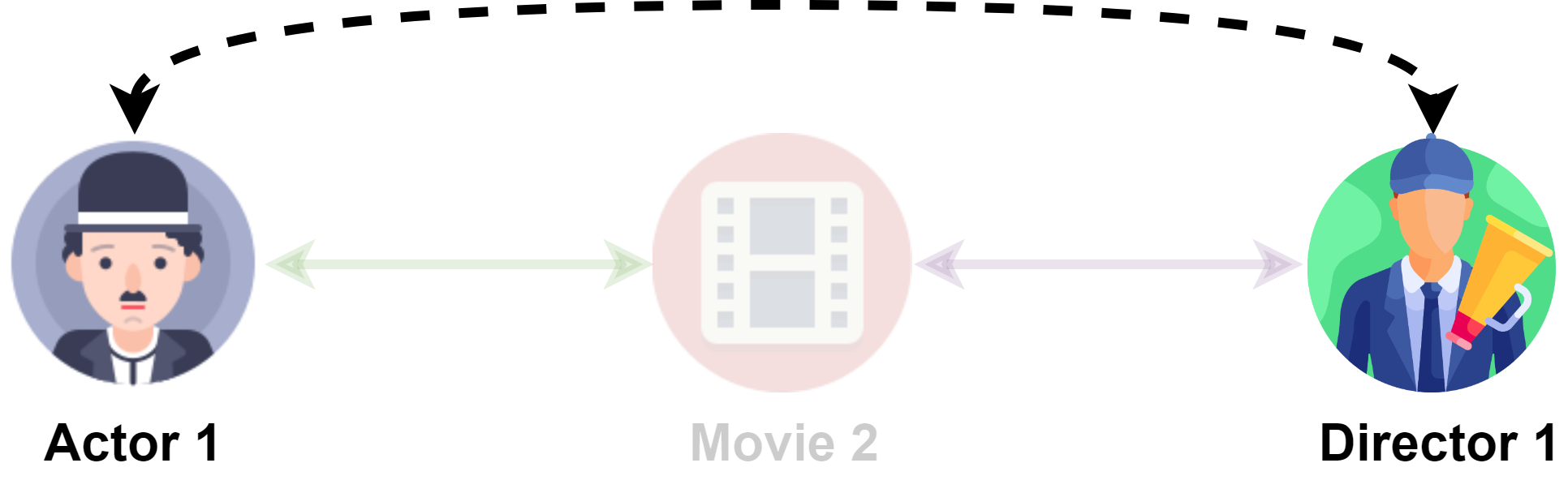}
        \caption{Metapath-guided Neighbor}
    \end{subfigure}
    \hfill
    \begin{subfigure}[b]{0.24\textwidth}
        \centering
        \includegraphics[width=\textwidth]{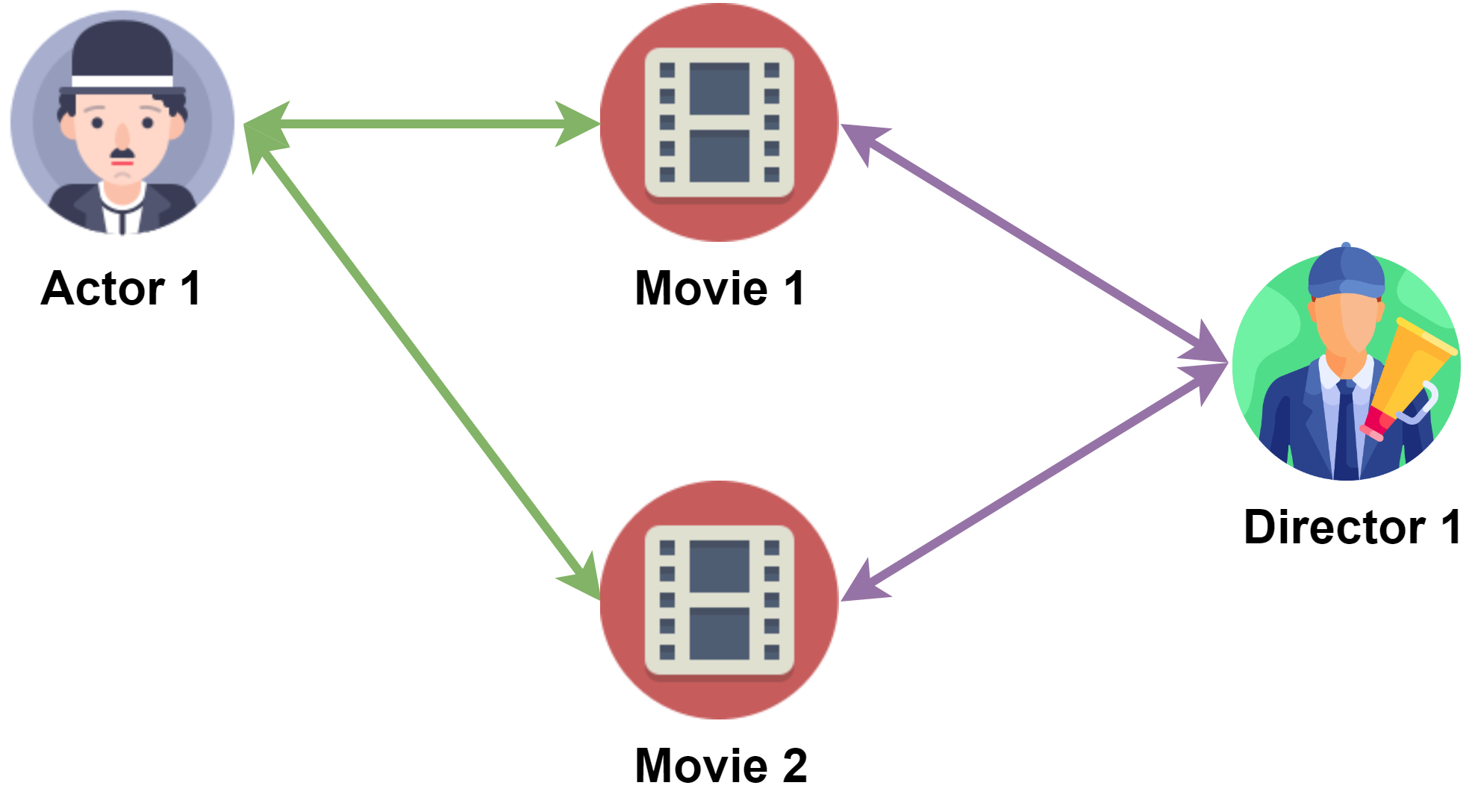}
        \caption{Metapath Context}
    \end{subfigure}
    \caption{An illustration of the terminologies defined in Section~\ref{sec:preliminary}.}
    \label{fig:illustration}
\end{figure*}

\subsection{Relation-based HGNNs}
Relation-based HGNNs aggregate direct neighbors modulated by type-specific weights.
By extending GCN with edge-type-specific weights, RGCN~\citep{DBLP:conf/esws/SchlichtkrullKB18} groups and aggregates the neighborhood based on the edge types.
{HetSANN~\citep{DBLP:conf/aaai/HongGLYLY20} employs type-specific graph attention layers to aggregate local information.}
HGT~\citep{DBLP:conf/www/HuDWS20} leverages a Transformer-like encoder~\citep{DBLP:conf/nips/VaswaniSPUJGKP17} parameterized by node and edge types to aggregate the local neighborhood.
Simple-HGN~\citep{DBLP:conf/kdd/LvDLCFHZJDT21} extends GAT to the heterogeneous setting via edge-type attention.
{R-HGNN~\citep{DBLP:journals/tkde/YuSDLLX23} jointly learns node and relation representations via relation-specific and cross-relation message passings.}
Many follow-up works are based on a similar idea~\citep{DBLP:conf/kdd/ZhangSHSC19, DBLP:journals/tkde/YangGLZCW23, DBLP:journals/corr/abs-2011-09679, DBLP:conf/nips/YoonPZHSP22, DBLP:conf/nips/AhnYGMW22}, which we do not describe in detail.
Despite simplicity and efficiency, the only way for relation-based HGNNs to capture information from nodes multiple hops away is to stack more GNN layers, bringing along potential performance degradation caused by the over-smoothing problem~\citep{DBLP:conf/aaai/LiHW18} and the over-use of transformation operations~\citep{DBLP:conf/cikm/ZhouDWLHXF21}.


\subsection{Metapath-based HGNNs}
To address the limitations stated above, some researchers designed HGNNs leveraging metapaths to associate nodes far from each other{, eliminating the need for stacking multiple layers}.
By connecting metapath-guided neighbors and discarding intermediate nodes, HAN~\citep{DBLP:conf/www/WangJSWYCY19} converts a heterogeneous graph into multiple homogeneous subgraphs. It then applies GAT to each subgraph and combines node representations from different metapaths using attention.
MAGNN~\citep{DBLP:conf/www/0004ZMK20} incorporates the intermediate nodes discarded by HAN through leveraging metapath instances. However, this raises an efficiency issue caused by duplicate nodes across an exploding number of metapath instances.
Metapaths in HAN and MAGNN are manually selected. To automate this process, GTN~\citep{DBLP:conf/nips/YunJKKK19} softly selects relations using learnable weights and applies matrix multiplication to obtain the adjacency matrix of arbitrary metapaths, which is fed to a GCN to get node representations. Useful metapaths can be automatically extracted based on the relation weights.
GTN requires a tremendous amount of memory space and computation time due to the large matrix multiplication step.
{Based on GTN, MEGNN~\citep{DBLP:journals/kbs/ChangCHZZC22} rearranges the calculation order of the matrix multiplications to improve the efficiency.}


Despite their success in different datasets, existing metapath-based HGNNs cannot have it all for performance and efficiency. HAN suffers from information loss. MAGNN and GTN are limited by high computation costs.
{MEGNN still faces the scalability issue with large graphs because of large matrix multiplication.}
Derivative studies based on these models share similar issues~\citep{DBLP:conf/aaai/ZhangWZSZZ22, DBLP:conf/www/00010YCLF022}.
{We leave the detailed illustration and discussion of the underlying mechanisms that cause these issues to Figure~\ref{fig:hetero_graph} and Section~\ref{sec:mgc}.}
Therefore, we want to design a metapath-based HGNN that can capture the most information from the input heterogeneous graph without information loss, and at the same time can be efficient without redundant computations.
We achieve these goals by defining and utilizing a special structure called \emph{metapath context}.
{The advantages of metapath context and how it can help avoid the issues mentioned above will be detailed in Section~\ref{sec:mcc}.}


\section{Preliminary} \label{sec:preliminary}
This section gives formal definitions of some basic terminologies of heterogeneous graphs.
Their graphical illustrations are provided in Figure~\ref{fig:illustration}.
Besides, Table~\ref{tab:notation} briefly references the mathematical symbols used in this work.

\begin{table}[t]
    \centering
    \caption{Notations used in this paper.}
    \label{tab:notation}
    \begin{tabular}{l|l}
        \toprule
        \textbf{Notation} & \textbf{Definition}\\
        \midrule
        $\mathbb{R}^{n}$ & $n$-dimensional Euclidean space\\
        $a$, $\mathbf{a}$, $\mathbf{A}$ & Scalar, vector, matrix\\
        $\mathbf{A}^{\intercal}$ & Matrix/vector transpose\\
        $\mathcal{G}=\left(\mathcal{V},\mathcal{E}\right)$ & A graph $\mathcal{G}$ with node set $\mathcal{V}$ and edge set $\mathcal{E}$\\
        $v$ & A node $v\in\mathcal{V}$\\
        $P$ & A metapath\\
        $\mathcal{P}_{A}$ & Metapaths starting from node type $A$\\
        $\mathcal{N}_v^{P}$ & Metapath-$P$-guided neighbors of node $v$\\
        $\mathbf{x}_v$ & Raw feature vector of node $v$\\
        $\mathbf{h}_v$ & Hidden states (embedding) of node $v$\\
        $\sigma(\cdot)$ & Activation function\\
        $\odot$ & Element-wise multiplication\\
        $|\cdot|$ & The cardinality of a set\\
        \bottomrule
    \end{tabular}
\end{table}

\begin{definition}[Heterogeneous Graph]
A heterogeneous graph is defined as a graph $\mathcal{G}=\left(\mathcal{V},\mathcal{E}\right)$ associated with a node type mapping function $\phi : \mathcal{V} \rightarrow \mathcal{A}$ and an edge type mapping function $\psi : \mathcal{E} \rightarrow \mathcal{R}$.
$\mathcal{A}$ and $\mathcal{R}$ denote the pre-defined sets of node types and edge types, respectively, with $|\mathcal{A}|+|\mathcal{R}|>2$.
\end{definition}

\begin{definition}[Metapath]
A metapath $P$ is defined as a path in the form of $A_{1} \stackrel{R_{1}}{\longrightarrow} A_{2} \stackrel{R_{2}}{\longrightarrow} \cdots \stackrel{R_{K}}{\longrightarrow} A_{K+1}$ (abbreviated as $A_{1} A_{2} \cdots A_{K+1}$), describing a composite relation $R=R_{1} \circ R_{2} \circ \cdots \circ R_{K}$ between node types $A_1$ and $A_{K+1}$, where $\circ$ denotes the composition operator on edge types.
\end{definition}

\begin{definition}[Metapath Instance]
Given a metapath $P=A_{1} \stackrel{R_{1}}{\longrightarrow} A_{2} \stackrel{R_{2}}{\longrightarrow} \cdots \stackrel{R_{K}}{\longrightarrow} A_{K+1}$ of a heterogeneous graph $G$, a metapath instance of $P$ is defined as a path $v_{1} \stackrel{e_{1}}{\longrightarrow} v_{2} \stackrel{e_{2}}{\longrightarrow} \cdots \stackrel{e_{K}}{\longrightarrow} v_{K+1}$ in graph $G$, such that $\phi\left(v_i\right)=A_i$ and $\psi\left(e_i\right)=R_i$ for all $i$.
\end{definition}

\begin{definition}[Metapath-guided Neighbor]
Given a metapath $P$ of a heterogeneous graph $\mathcal{G}$, the metapath-guided neighbors $\mathcal{N}_v^{P}$ of a node $v$ is defined as the set of nodes that connect with node $v$ via some metapath instances of $P$.
\end{definition}

\begin{definition}[Metapath Context] \label{def:mc}
Given a metapath $P$ of a heterogeneous graph $\mathcal{G}$, the metapath context of a node $v$ is the local subgraph $\mathcal{S}_v^P = \left(\mathcal{V}_v^P, \mathcal{E}_v^P\right)$ centered at $v$, such that edge $(v_1, v_2) \in \mathcal{E}_v^P$ if and only if it belongs to a metapath instance of $P$ starting from $v$ in the original graph $\mathcal{G}$.
\end{definition}

\begin{definition}[Heterogeneous Graph Representation Learning]
Given a heterogeneous graph $\mathcal{G}=\left(\mathcal{V},\mathcal{E}\right)$, with node feature matrices $\mathbf{X}_{A_i} \in \mathbb{R}^{|\mathcal{V}_{A_i}| \times d_{A_i}}$ for node types $A_i \in \mathcal{A}$, heterogeneous graph representation learning is to learn the $d$-dimensional node vectors $\mathbf{h}_{v} \in \mathbb{R}^{d}$ for all $v \in \mathcal{V}$ with $d \ll |\mathcal{V}|$ that are able to capture rich structural and semantic information involved in $\mathcal{G}$.
\end{definition}

\section{Methodology} \label{sec:methodology}

In this section, we propose MECCH, a novel metapath context convolution-based HGNN.
First in Section~\ref{sec:framework}, we describe a general framework that can represent most existing metapath-based HGNNs.
Then in Section~\ref{sec:mcc}, \ref{sec:mce}, and \ref{sec:conv_mf}, we present the three novel components that constitute MECCH.
In the end, we also introduce the training losses in Section~\ref{sec:loss} and analyze the model complexity in Section~\ref{sec:complexity}.

\subsection{Metapath-based HGNN Framework} \label{sec:framework}

We formalize a metapath-based HGNN as the following four components applied in series: (1) \emph{feature preprocessing}, (2) \emph{local subgraph re-construction (LSR)}, (3) \emph{metapath-specific graph convolution (MGC)}, and (4) \emph{metapath fusion (MF)}.
Existing metapath-based HGNNs differ in the last three components.
We summarize them along with MECCH in terms of the LSR, MGC, and MF components in Table~\ref{tab:framework}, where MN denotes metapath-guided neighbors, MI denotes metapath instances, and MC denotes metapath contexts.

\begin{table}[t]
\centering
\caption{A summary of metapath-based HGNNs following our framework.}
\label{tab:framework}
\begin{tabular}{@{}cccc@{}}
\toprule
\textbf{Model} & \textbf{LSR}   & \textbf{MGC}             & \textbf{MF}           \\ \midrule
HAN   & MN    & GAT             & Attention    \\
MAGNN & MI    & GAT+RotatE      & Attention    \\
GTN   & MN    & GCN             & Weighted Sum \\
MECCH & MC    & Graph Pooling   & 1-D Convolution     \\ \bottomrule
\end{tabular}
\end{table}

\subsubsection{Feature Preprocessing} In heterogeneous graphs, different node types may have unequal dimensions of node features. Therefore, before feeding the input graph into an HGNN, one would apply a node-type-specific linear transformation to project node features into latent vectors with the same dimensionality. For nodes without raw features, we can assign one-hot vectors to them as initial features. Combined with the linear transformation, this is equivalent to an embedding lookup. The projected representation $h_v^0$ of node $v$ of type $\phi\left(v\right)$ can be obtained by:
\begin{equation}
\label{eq:feat_pre}
    \mathbf{h}_{v}^{0} = \mathbf{W}_{\phi\left(v\right)} \mathbf{x}_{v} + \mathbf{b}_{\phi\left(v\right)},
\end{equation}
where $\mathbf{x}_{v} \in \mathbb{R}^{d_{\phi\left(v\right)}}$ is the raw feature, $\mathbf{W}_{\phi\left(v\right)} \in \mathbb{R}^{d \times d_{\phi\left(v\right)}}$ and $\mathbf{b}_{\phi\left(v\right)} \in \mathbb{R}^{d}$ are the learnable weights and bias.

\subsubsection{Local Subgraph Re-construction}
In metapath-based HGNNs, nodes from the input graph need to be re-connected for message passing through metapath instead of from direct neighbors.
Following this, a new local subgraph $\mathcal{S}_v^P = \left(\mathcal{V}_v^P, \mathcal{E}_v^P\right)$ would be re-constructed for each node in each related metapath. The definition of $\mathcal{S}_v^P$ differs across HGNNs. For example, $\mathcal{S}_v^P$ in HAN is the metapath-$P$-guided neighborhood of $v$, while $\mathcal{S}_v^P$ in MAGNN is the set of metapath-$P$'s instances incident to $v$. Each $\mathcal{S}_v^P$ carries unique semantics that will be encoded and captured in the next step.


\begin{figure}[tb]
    \centering
    \includegraphics[width=0.9\columnwidth]{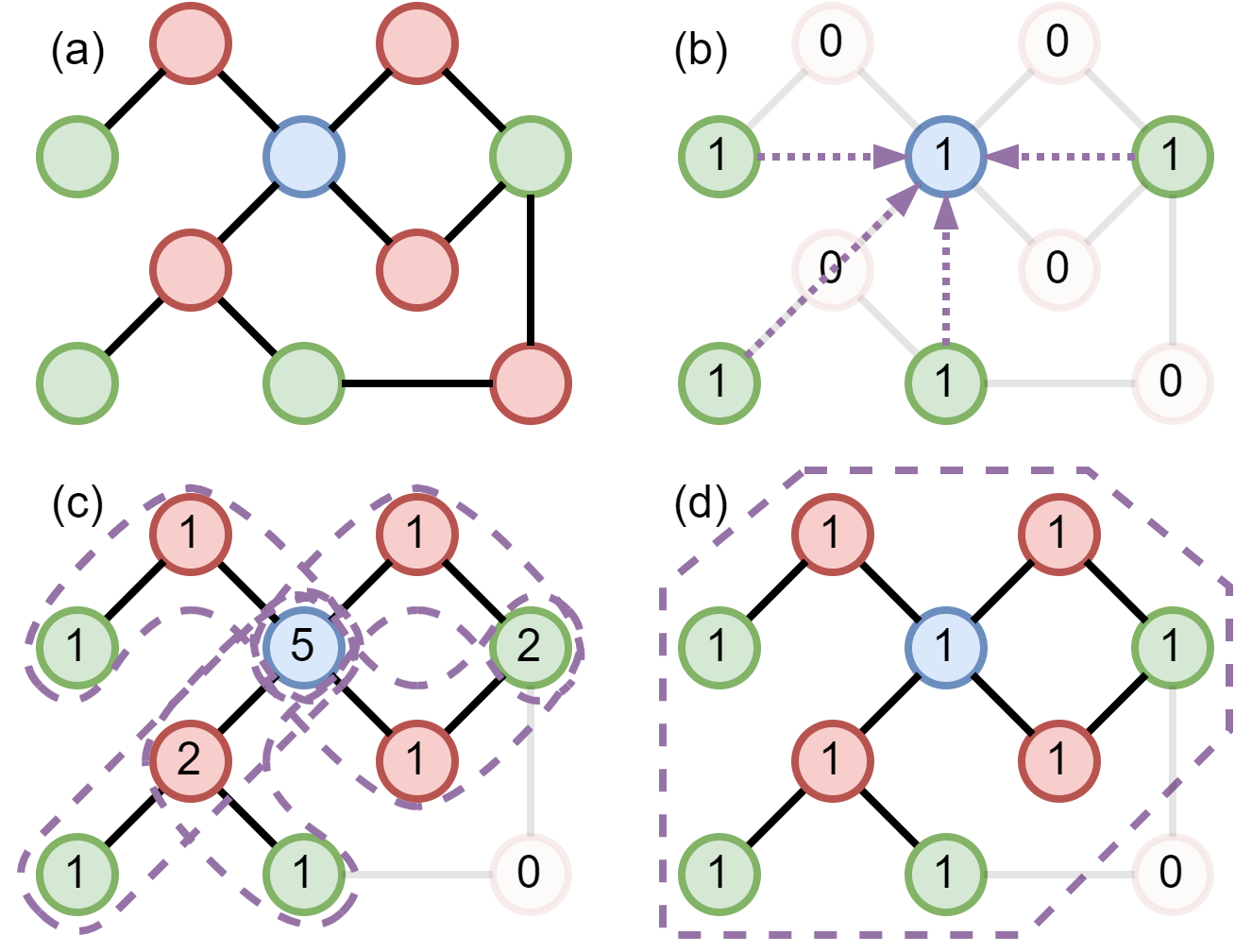}
    \caption{HGNN aggregation for the blue target node using Blue-Red-Green metapath. Numbers indicate how many times the nodes are aggregated. (a) A sample heterogeneous graph. (b) HAN \& GTN: metapath-guided neighbors. (c) MAGNN: metapath instances. (d) MECCH (ours): metapath context.}
    \label{fig:hetero_graph}
\end{figure}

\subsubsection{Metapath-specific Graph Convolution} \label{sec:mgc}
To compute metapath-$P$-specific node representations, the model would conduct graph convolution on the re-constructed subgraphs $\mathcal{S}_v^P$ from the last step.
For node $v$ at layer $l$, the metapath-$P$-specific node representation $\mathbf{h}_{v,P}^l$ is computed by:
\begin{equation}
    \mathbf{h}_{v,P}^l = \operatorname{MGC}_P^l\left(\mathcal{S}_v^P, \left\{\mathbf{h}_u^{l-1}, \forall u \in \mathcal{V}_v^P \right\}\right),
\end{equation}
where $\operatorname{MGC}_P^l$ is the graph convolution function for metapath $P$ at layer $l$.
Different models could have different designs for this function. For example, HAN ignores the intermediate nodes and directly applies GAT to aggregate metapath-guided neighbors. MAGNN incorporates intermediate nodes by encoding and aggregating metapath instances. GTN essentially applies GCN to the metapath-guided neighbors for automatically generated metapaths. A graphical illustration of different $\operatorname{MGC}_P^l$ implementations is provided in Figure~\ref{fig:hetero_graph}.
{Basically, the existing metapath-based HGNNs either aggregate deficient nodes (which causes information loss), or aggregate redundant nodes (which leads to high computation costs). Our MECCH introduced later aims to extract adequate information from the graph without any redundancy.}
After applying this component, every node is now associated with a set of vector representations, each from one metapath.

\begin{figure*}[t]
    \centering
    \includegraphics[width=0.95\textwidth]{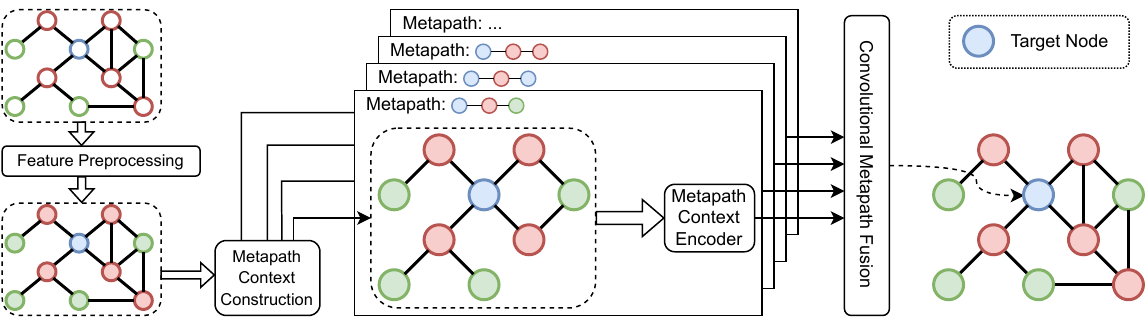}
    \caption{The overall architecture of MECCH. To obtain the representation of the blue target node, after feature preprocessing, MECCH \emph{constructs the metapath contexts} using metapaths starting from the blue node type. Then, MECCH \emph{encodes each metapath context} into a vector representation. Finally, MECCH \emph{fuses the context representations} into one node representation via 1-D convolution.}
    \label{fig:model}
\end{figure*}

\subsubsection{Metapath Fusion}
To combine unilateral metapath-specific representations into one unified node representation $\mathbf{h}_{v}^l$, metapath-based HGNNs require a fusion step $\operatorname{MF}_{\phi\left(v\right)}^l$:
\begin{equation}
    \mathbf{h}_{v}^l = \operatorname{MF}_{\phi\left(v\right)}^l\left(\left\{\mathbf{h}_{v,P}^l, \forall P \in \mathcal{P}_{\phi\left(v\right)}\right\}\right),
\end{equation}
where $\mathcal{P}_{\phi\left(v\right)}$ is the set of metapaths starting from node type $\phi\left(v\right)$, either manually selected or automatically generated.

Our proposed MECCH follows this framework with novel designs on the last three components. Specifically, after feature preprocessing, we propose to construct \emph{metapath contexts} for each node in each metapath, then utilize a \emph{metapath context encoder} to embed metapath-specific information, and finally leverage \emph{convolutional metapath fusion} to integrate information from different metapaths together.
Figure~\ref{fig:model} illustrates the overall architecture of MECCH.
The model pseudocode is provided in Algorithm~\ref{alg:algorithm}.

\subsection{Metapath Context Construction} \label{sec:mcc}

After feature preprocessing, for each node $v$ in each metapath $P \in \mathcal{P}_{\phi\left(v\right)}$, MECCH constructs a metapath context, a local subgraph $\mathcal{S}_v^P$ centered at $v$ based on Definition~\ref{def:mc}.
The metapath context $\mathcal{S}_v^P$ describes the entire structure connecting the target node $v$ and the metapath-$P$-guided neighbors.
This notion avoids information loss caused by discarding intermediate nodes in HAN, and reduces redundant computations caused by duplicate nodes across metapath instances in MAGNN.
The concept of metapath context is illustrated in Figure~\ref{fig:hetero_graph}.
{As shown, by using metapath contexts, MECCH aggregates each node within the receptive field exactly once, thus can achieve improved efficiency with optimal performance.}

For the metapaths considered, MECCH uses all metapaths of a pre-defined length $K$, which does not require a manual selection process based on human expertise.
Any metapath of length less than $K$ must be a sub-sequence of some length-$K$ metapath. The shorter metapath's context would be subsumed entirely as a subgraph of a longer metapath's context.
Considering any metapath shorter than $K$ would lead to extracting redundant information.
Therefore, unlike GTN, which uses all metapaths with length $\leq K$, MECCH only considers length-$K$ metapaths.
In this way, MECCH requires no prior knowledge to select metapaths and at the same time avoids a significant amount of redundant computations.

\subsection{Metapath Context Encoder} \label{sec:mce}

After constructing the metapath context $\mathcal{S}_v^P$, MECCH employs a metapath context encoder to learn structural and semantic information embedded in $\mathcal{S}_v^P$. Since metapath contexts are essentially graphs, it is natural to leverage graph pooling functions to encode $\mathcal{S}_v^P$. Therefore, the metapath context encoder $\operatorname{MGC}_P^l$ of MECCH is formulated as:
\begin{equation}
\label{eq:mce}
    \mathbf{h}_{v,P}^l = \operatorname{READOUT}_P^l\left(\mathcal{S}_v^P,\left\{\mathbf{h}_u^{l-1},\forall u \in \mathcal{S}_v^P\right\}\right),
\end{equation}
where $\operatorname{READOUT}_P^l$ could be a permutation-invariant function like summation, or a more sophisticated GNN-based graph encoder to consider the graph structure.
Each $\mathbf{h}_{v,P}^l$ obtained can be interpreted as the information of node $v$ exhibited by metapath $P$.
Our experiments show that an attention-based encoder cannot well capture the semantics embedded in the metapath context.
Eventually, we employ the mean graph pooling strategy for this component, which is highly efficient without the need to tune any parameters.

\subsection{Convolutional Metapath Fusion} \label{sec:conv_mf}

After encoding metapath contexts, MECCH applies a fusion layer to combine representations from different metapaths $\left\{\mathbf{h}_{v,P}^l, \forall P \in \mathcal{P}_{\phi\left(v\right)}\right\}$ into one unified node representation.
A straightforward solution is to take the element-wise mean. But the mean operation does not consider that metapaths might have different importance to node representations.
HAN and MAGNN compute metapath weights by applying attention to metapath summary vectors calculated by averaging projected representations of all nodes, which restricts the model to full-batch training only.
To bypass this limitation, MECCH adopts an efficient \mbox{1-D} convolution kernel to adaptively learn the impact of each metapath on the downstream task:
\begin{equation}
\label{eq:conv_mf_1}
    \mathbf{h}_{v,\mathcal{P}_{\phi\left(v\right)}}^l = \sum_{P\in \mathcal{P}_{\phi\left(v\right)}} \boldsymbol{\alpha}_P^l \odot \mathbf{h}_{v,P}^l,
\end{equation}
where $\boldsymbol{\alpha}_P^l$ is a learnable vector of length equal to the hidden dimension $d$. Then MECCH further applies a projection to obtain the final node vector of the desired output dimension:
\begin{equation}
\label{eq:conv_mf_2}
    \mathbf{h}_v^l = \sigma\left(\mathbf{W}_{\phi\left(v\right)}^l\mathbf{h}_{v,\mathcal{P}_{\phi\left(v\right)}}^l + \mathbf{b}_{\phi\left(v\right)}^l\right),
\end{equation}
where $\mathbf{W}_{\phi\left(v\right)}^l$ and $\mathbf{b}_{\phi\left(v\right)}^l$ are the weight matrix and the bias term, $\sigma$ is an activation function (ReLU by default).

\begin{algorithm}[t]
\caption{MECCH forward propagation.} \label{alg:algorithm}
\textbf{Input}: Heterogeneous graph $\mathcal{G}=\left(\mathcal{V},\mathcal{E}\right)$ with node type mapping $\phi : \mathcal{V} \rightarrow \mathcal{A}$ and edge type mapping $\psi : \mathcal{E} \rightarrow \mathcal{R}$; Node features $\left\{\mathbf{x}_{v}, \forall v \in \mathcal{V}\right\}$; Metapath sets $\left\{\mathcal{P}_{A}, \forall A \in \mathcal{A}\right\}$; Number of layers $L$\\
\textbf{Output}: Node representations
\begin{algorithmic}[1] 
\State $\mathbf{h}_{v}^{0} = \mathbf{W}_{\phi\left(v\right)} \mathbf{x}_{v} + \mathbf{b}_{\phi\left(v\right)}, \forall v \in \mathcal{V}$ \Comment{Feature preprocessing}
\ForAll{$A \in \mathcal{A}$}
    \ForAll{$P \in \mathcal{P}_A$}
        \State Construct $\mathcal{S}_v^P = \left(\mathcal{V}_v^P, \mathcal{E}_v^P\right), \forall v \in \mathcal{V}_A$ \Comment{\ref{sec:mcc}}
    \EndFor
\EndFor
\For{$l = 1 \ldots L$}
    \ForAll{$A \in \mathcal{A}$}
        \ForAll{$P \in \mathcal{P}_A$}
            \State $\mathbf{h}_{v,P}^l = \operatorname{READOUT}_P^l\left(\mathcal{S}_v^P, \left\{\mathbf{h}_u^{l-1},\forall u \in \mathcal{S}_v^P\right\}\right),$
            \Statex[3] $\forall v \in \mathcal{V}_A$ \Comment{\ref{sec:mce}}
        \EndFor
        \State $\mathbf{h}_{v,\mathcal{P}_{A}}^l = \sum_{P\in \mathcal{P}_{A}} \boldsymbol{\alpha}_P^l \odot \mathbf{h}_{v,P}^l, \forall v \in \mathcal{V}_A$ \Comment{\ref{sec:conv_mf}}
        \State $\mathbf{h}_v^l = \sigma\left(\mathbf{W}_{A}^l\mathbf{h}_{v,\mathcal{P}_{A}}^l + \mathbf{b}_{A}^l\right), \forall v \in \mathcal{V}_A$ \Comment{\ref{sec:conv_mf}}
    \EndFor
\EndFor
\State \textbf{return} $\left\{\mathbf{h}_{v}^L, \forall v \in \mathcal{V}\right\}$
\end{algorithmic}
\end{algorithm}

\subsection{Training Losses} \label{sec:loss}

For the node classification task, the loss function would be the cross-entropy between ground truths and model outputs over a small set of labeled nodes:
\begin{equation}
    \mathcal{L}_{nc}=-\frac{1}{|\mathcal{V}_{label}|}\sum_{v \in \mathcal{V}_{label}} \sum_{c=1}^{C} \mathbf{y}_{v}[c] \cdot \log \hat{\mathbf{y}}_{v}[c],
\end{equation}
where $\mathcal{V}_{label}$ is the set of labeled nodes for training, $C$ is the number of classes, $\mathbf{y}_{v}$ is the one-hot ground truth of node $v$, and $\hat{\mathbf{y}}_{v} = \mathbf{h}_v^L$ is the model's predicted probability vector.

For the link prediction task, we leverage the negative sampling strategy~\citep{DBLP:conf/nips/MikolovSCCD13} to train the model along with a DistMult~\citep{DBLP:journals/corr/YangYHGD14a} link decoder using a binary cross-entropy loss function:
\begin{equation}
\begin{split}
    \mathcal{L}_{lp} = -\frac{1}{|\Omega|}\sum_{\left(u,v\right) \in \Omega} \Big(&\log \sigma\left(\mathbf{h}_{u}^{\intercal} \mathbf{W}_{lp} \mathbf{h}_{v} \right) +\\
    &\mathbb{E}_{v^- \sim p_n}\left[\log \sigma\left(-\mathbf{h}_{u}^{\intercal} \mathbf{W}_{lp} \mathbf{h}_{v^-} \right)\right]\Big),
\end{split}
\end{equation}
where $\mathbf{W}_{lp}=\operatorname{diag}(\mathbf{w}_{lp})$ is a learnable diagonal matrix, $\sigma$ is the sigmoid function, $\Omega$ is the set of positive (observed) edges, and $p_n$ is the distribution of $v^-$ for sampling negative edges $\left(u, v^-\right)$.

\subsection{Complexity Analysis} \label{sec:complexity}


The time complexity of MECCH consists of three parts, corresponding to the feature preprocessing, the metapath context encoder, and the convolutional metapath fusion, respectively.
The metapath context construction does not contribute to the complexity because it can be pre-computed.
To simplify the complexity analysis, we consider the following scenario: there are $N$ nodes in the input graph, every node is associated with $M$ metapaths of length $K$, and every node has $t$ type-$A$ neighbors for each node type $A$.
The feature preprocessing is done by Eq.~(\ref{eq:feat_pre}), which requires $\mathcal{O}(N_A \cdot d \cdot d_A)$ for each node type $A$. If we assume every node type's input features have the same dimensionality $d_{in}$. Then this part requires $\mathcal{O}(N \cdot d \cdot d_{in})$ in total.
The metapath context encoder is defined by Eq.~(\ref{eq:mce}), where $\mathcal{S}_v^P$ contains at most $1+t+t^2+\cdots+t^K=\frac{t^{K+1}-1}{t-1}$ nodes. Hence this part requires $\mathcal{O}(N \cdot M \cdot \frac{t^{K+1}-1}{t-1} \cdot d) = \mathcal{O}(N \cdot M \cdot t^K \cdot d)$.
The convolutional metapath fusion consists of Eq.~(\ref{eq:conv_mf_1}) and Eq.~(\ref{eq:conv_mf_2}). Eq.~(\ref{eq:conv_mf_1}) requires $\mathcal{O}(N \cdot M \cdot d)$, while Eq.~(\ref{eq:conv_mf_2}) requires $\mathcal{O}(N \cdot d \cdot d_{out})$.

In practice, the input dimensionality $d_{in}$ and the output dimensionality $d_{out}$ are typically very small compared to $M \cdot t^K$. Thus, the overall time complexity of a 1-layer MECCH is $\mathcal{O}(N \cdot M \cdot t^K \cdot d)$.
In comparison, MAGNN requires $\mathcal{O}(N \cdot M \cdot K \cdot t^K \cdot d)$, GTN requires $\mathcal{O}(N^3 \cdot K)$, and HAN requires $\mathcal{O}(N \cdot M \cdot t^K \cdot d)$. Therefore, we have MECCH $\approx$ HAN $<$ MAGNN $<$ GTN for computational complexity.






\begin{table*}[t]
\centering
\caption{Statistics of datasets.}
\label{tab:dataset}
\begin{tabular}{@{}crrrcrrr@{}}
\toprule
\textbf{Dataset} & \multicolumn{1}{c}{\textbf{\# Nodes (Types)}} & \multicolumn{1}{c}{\textbf{\# Edges (Types)}} & \multicolumn{1}{c}{\textbf{\# Feats}} & \textbf{Target} & \multicolumn{1}{c}{\textbf{\# Training}} & \multicolumn{1}{c}{\textbf{\# Validation}} & \multicolumn{1}{c}{\textbf{\# Testing}} \\ \midrule
IMDB    & 12,722 (3)             & 37,288 (4)             & 1,256                           & Movie  & 300 (10\%)                             & 300 (10\%)                               & 2,339 (80\%)                          \\
ACM     & 8,994 (3)             & 25,922 (4)             & 1,902                           & Paper  & 600 (20\%)                             & 300 (10\%)                               & 2,125 (70\%)                          \\
DBLP    & 18,405 (3)             & 67,946 (4)             & 334                             & Author & 800 (20\%)                             & 400 (10\%)                               & 2,857 (70\%)                          \\
LastFM   & 20,612 (3)             & 201,908 (5)             & -                             & User-Artist & 64,984 (70\%)                             & 9,283 (10\%)                               & 18,567 (20\%)                          \\
PubMed   & 63,109 (4)             & 368,245 (16)             & 200                             & Disease-Disease & 29,845 (70\%)                             & 4,264 (10\%)                               & 8,528 (20\%)                          \\ \bottomrule
\end{tabular}
\end{table*}

\section{Experiments} \label{sec:experiments}

In this section, we demonstrate the effectiveness and efficiency of our proposed MECCH by conducting experiments in the node classification and link prediction tasks on five heterogeneous graph datasets.
The experiments are designed to answer the following research questions.
\begin{itemize}
    \item RQ1. How does MECCH perform in classifying nodes?
    \item RQ2. How does MECCH perform in predicting links?
    \item RQ3. How efficient is MECCH?
    \item RQ4. How effective is each proposed component?
\end{itemize}

\subsection{Experimental Setups}

\subsubsection{Datasets}
We select datasets of varying sizes from different domains to simulate real-life applications.
For \emph{node classification}, we adopt the IMDB, ACM, and DBLP datasets from \citep{DBLP:conf/nips/YunJKKK19} to evaluate the performance of MECCH and baselines.
For \emph{link prediction}, we choose the LastFM dataset from \citep{DBLP:conf/www/0004ZMK20} and the PubMed dataset from \citep{DBLP:journals/tkde/YangXZSH22} to compare the models.
Since previous researchers usually preprocess the raw data under different pipelines, scores reported in their original papers cannot be compared directly. To ensure fairness, we reran experiments of each model on the same preprocessed datasets.
{We try our best to follow the default settings (e.g., training/validation/testing splits) of these datasets.}
Statistics of the five adopted datasets are summarized in Table~\ref{tab:dataset}.
\begin{itemize}
    \item \textbf{IMDB} is a movie information network for \emph{node classification}. The preprocessed subset contains three types of nodes: movies, actors, and directors. A bag-of-words representation of plot keywords describes each node in the dataset. The movie nodes are labeled by their genres.
    \item \textbf{ACM} is an academic network for \emph{node classification}, with three types of nodes: papers, authors, and subjects. Nodes in ACM are associated with bag-of-words representations of keywords. The paper nodes are labeled based on the conference in which they were published.
    \item \textbf{DBLP} is also an academic network for \emph{node classification}. It contains three types of nodes: papers, authors, and conferences. Nodes in the dataset are associated with bag-of-words representations of keywords. The author nodes are labeled based on their research fields.
    \item \textbf{LastFm} is a music listening information network for \emph{link prediction}. The preprocessed subset contains three types of nodes: users, artists, and tags. Nodes in this dataset are not associated with any features. The prediction targets are the user-artist edges.
    \item \textbf{PubMed} is a biomedical knowledge graph for \emph{link prediction}, with four types of nodes: genes, diseases, chemicals, and species. Nodes in PubMed are associated with aggregated word2vec embeddings. The prediction targets are the disease-disease edges.
\end{itemize}

\subsubsection{Baselines}
We compare MECCH against state-of-the-art HGNNs, including three \emph{relation-based HGNNs} (RGCN, HGT, and Simple-HGN), three \emph{metapath-based HGNNs} (HAN, MAGNN, and GTN).
\begin{itemize}
    \item \textbf{RGCN}~\citep{DBLP:conf/esws/SchlichtkrullKB18} is a \emph{relation-based HGNN} that groups and aggregates neighbors based on their edge types with edge-type-specific weights.
    \item \textbf{HGT}~\citep{DBLP:conf/www/HuDWS20} is a \emph{relation-based HGNN}. It leverages a Transformer-like self-attention architecture parameterized by node types and edge types to capture the importance weights of the heterogeneous neighborhood.
    \item \textbf{Simple-HGN}~\citep{DBLP:conf/kdd/LvDLCFHZJDT21} is a \emph{relation-based HGNN}. It extends GAT by adding edge-type attention, residual connection, and output normalization.
    \item \textbf{HAN}~\citep{DBLP:conf/www/WangJSWYCY19} is a \emph{metapath-based HGNN}. It learns metapath-specific node embeddings by aggregating metapath-guided neighbors and leverages the attention mechanism to combine metapaths.
    \item \textbf{MAGNN}~\citep{DBLP:conf/www/0004ZMK20} is a \emph{metapath-based HGNN} that enhances HAN by incorporating the intermediate nodes along metapaths with metapath instance encoders.
    \item \textbf{GTN}~\citep{DBLP:conf/nips/YunJKKK19} is a \emph{metapath-based HGNN} that can automatically discover valuable metapaths via multiplying weighted sums of relation subgraphs.
\end{itemize}

\subsubsection{Hyperparameters}
For fairness, we configure the node embedding dimension of all HGNNs to 64.
For models with multi-head attention, we set the number of heads as 8 and ensure it would not multiply the embedding dimension, as suggested in \citep{DBLP:conf/www/HuDWS20}.
For metapaths used by HAN and MAGNN, we adopt the ones suggested in their papers.
For MAGNN, we use the relational rotation metapath instance encoder.
For GTN, we use the adaptive learning rate as suggested in the authors' implementation.
For MECCH, we choose the metapath length $K$ within $\{1, 2, 3, 4, 5\}$.
For each model-dataset pair, we experiment with the number of HGNN layers within $\{1, 2, 3, 4, 5\}$, the dropout rate within $\{0, 0.5\}$, the learning rate within $\{1, 2, 5\} \times\left\{10^{-3}, 10^{-2}\right\}$, the weight decay within $\left\{0, 10^{-3}\right\}$.
The hyperparameter combinations that can fit in the GPU memory and yields the best results are selected (except for GTN, which runs on CPU for ACM, DBLP, and LastFM).
We use the same data splits for training, validation, and testing across models.
We train all HGNNs using the Adam optimizer for 500 epochs with early stopping applied after a patience of 50 epochs.

\begin{table*}[t]
\centering
\caption{Experiment results (\%) of the node classification task on the IMDB, ACM, and DBLP datasets. $\uparrow$ ($\downarrow$) indicates the higher (lower), the better. For each column, the best result is in \textbf{bold} font, and the second-best result is \underline{underlined}.}
\label{tab:experiment_node}
\begin{tabular}{@{}ccccccc@{}}
\toprule
\multirow{2}{*}{\textbf{Model}} & \multicolumn{2}{c}{\textbf{IMDB}}  & \multicolumn{2}{c}{\textbf{ACM}}   & \multicolumn{2}{c}{\textbf{DBLP}} \\ \cmidrule(l){2-7}
                      & Macro-F1$\uparrow$    & Micro-F1$\uparrow$    & Macro-F1$\uparrow$    & Micro-F1$\uparrow$    & Macro-F1$\uparrow$    & Micro-F1$\uparrow$    \\ \midrule
RGCN                   & 56.85$\pm$0.41 & 58.59$\pm$0.42 & \underline{92.17$\pm$0.28} & \underline{92.07$\pm$0.27} & 92.46$\pm$1.40 & 93.41$\pm$1.17 \\
HGT                    & 57.32$\pm$1.04 & 59.02$\pm$1.33 & 90.55$\pm$0.63 & 90.52$\pm$0.63 & 93.96$\pm$0.24 & 94.76$\pm$0.24 \\
Simple-HGN             & \underline{58.70$\pm$0.63} & 60.27$\pm$0.85 & 89.36$\pm$3.12 & 89.25$\pm$3.17 & 92.60$\pm$0.53 & 93.40$\pm$0.52 \\
HAN                    & 58.54$\pm$1.67 & 59.60$\pm$1.80 & 91.67$\pm$0.18 & 91.55$\pm$0.18 & 92.29$\pm$0.35 & 93.16$\pm$0.34 \\
MAGNN                  & 57.87$\pm$1.29 & 59.78$\pm$1.65 & 90.63$\pm$0.27 & 90.58$\pm$0.26 & \underline{94.14$\pm$0.23} & \underline{94.86$\pm$0.23} \\
GTN                    & 58.66$\pm$1.30 & \underline{60.28$\pm$1.74} & 91.94$\pm$0.57 & 91.83$\pm$0.57 & 93.25$\pm$0.25 & 94.12$\pm$0.21 \\ \midrule
MECCH                  & \textbf{62.59$\pm$1.96} & \textbf{64.62$\pm$2.38} & \textbf{92.74$\pm$0.40} & \textbf{92.67$\pm$0.41} & \textbf{94.34$\pm$0.29} & \textbf{95.08$\pm$0.25} \\ \bottomrule
\end{tabular}
\end{table*}

\subsubsection{Platform Specifications}


All of our experiments are conducted on the same platform with the following system specifications. \textbf{CPU}: 40x Intel Xeon Platinum 8268 cores. \textbf{GPU}: 1x NVIDIA TITAN V card. \textbf{Memory}: 775 GiB. \textbf{Operating System}: Ubuntu 18.04.6 LTS. \textbf{Libraries}: DGL 0.7.2, PyTorch 1.10.1, and CUDA 11.3.

\begin{table}[t]
\centering
\caption{Experiment results (\%) of the link prediction task on the LastFM and PubMed datasets.} 
\label{tab:experiment_link}
\begin{tabular}{@{}ccc@{}}
\toprule
\textbf{Model} & \textbf{LastFM}  & \textbf{PubMed} \\ \midrule
RGCN                   & 82.85$\pm$0.61 & 68.22$\pm$1.26 \\
HGT                    & 82.36$\pm$0.29 & 73.24$\pm$1.46 \\
Simple-HGN             & \underline{83.02$\pm$0.91} & \underline{73.74$\pm$0.50} \\
HAN                    & 82.26$\pm$0.42 & 71.50$\pm$0.84 \\
MAGNN                  & 79.26$\pm$0.89 & 65.74$\pm$2.35 \\
GTN                    & 76.94$\pm$0.09 & - \\ \midrule
MECCH                  & \textbf{84.27$\pm$0.05} & \textbf{76.69$\pm$0.24} \\ \bottomrule
\end{tabular}
\end{table}

\subsection{Node Classification (RQ1)}

We evaluate MECCH against other baselines for node classification on IMDB, ACM, and DBLP.
As shown in Table~\ref{tab:dataset}, only a small set of labeled nodes are available for training and validation, which is a typical semi-supervised node classification setting for GNNs.
The averages and standard deviations of \emph{Macro-F1} and \emph{Micro-F1} scores from 5 runs of each model on each dataset are reported in Table~\ref{tab:experiment_node}. From the table, we have the following observations:
\begin{enumerate}
    \item MECCH consistently outperforms previous state-of-the-art HGNNs in node classification across different datasets. On average, the performance gain of our model over the best baseline is around 2.59\%.
    \item Generally speaking, metapath-based HGNNs produce higher and more stable results than relation-based HGNNs. But the margin is tiny. Nevertheless, our MECCH proves that properly leveraging metapath contexts can bring noticeable advantages.
    \item It is interesting that MAGNN and GTN can hardly benefit from their high computation costs. One possible reason might be that redundant aggregations overly smooth meaningful signals from important nodes. {For MAGNN, another possible culprit is the parameter-intensive attention mechanism, which is removed in MECCH to avoid overfitting and enhance generalization ability.}
\end{enumerate}


\subsection{Link Prediction (RQ2)}

On LastFM and PubMed, link prediction is formulated as a binary classification problem to distinguish whether a given edge exists or not in the original graph.
During training, the negative edge $\left(u, v^-\right)$ is generated by replacing the destination node $v$ of a positive edge $\left(u, v\right)$ by uniformly sampling $v^-$ from $\mathcal{V}_{\phi\left(v\right)}$.
For validation and testing, negative edges are pre-generated in a 1:1 ratio to the positive ones and are kept the same across different models for a fair comparison.
We also follow \citep{DBLP:conf/kdd/LvDLCFHZJDT21} to sample hard negative edges for validation and testing.
The averages and standard deviations of \emph{ROC-AUC} scores in 5 runs are reported in Table~\ref{tab:experiment_link}, from which we have the following observations:
\begin{enumerate}
    \item MECCH performs consistently better than other models, with 1.51\% and 4.00\% improvements over the strongest baselines on LastFM and PubMed, respectively.
    \item The small ROC-AUC variances show that MECCH is much more stable than other HGNNs.
    \item One interesting observation is that relation-based HGNNs usually achieve better scores than metapath-based HGNNs in the link prediction task. Nevertheless, our MECCH still outperforms all of them.
    \item On PubMed, MAGNN has poor performance, and GTN cannot run due to significant memory consumption. They could both suffer from the diluted neighborhood messages caused by aggregating duplicate information. {Moreover, MAGNN may also be negatively affected by the over-parameterized attention modules.}
\end{enumerate}
\begin{table}[t]
\centering
\caption{Efficiency comparisons of metapath-based HGNNs on ACM and DBLP. TPE: time per epoch (s); {\#E: number of epochs;} RAM: CPU memory usage (GiB); VRAM: GPU memory usage (GiB).}
\label{tab:efficiency}
\begin{tabular}{@{}c|crrrr@{}}
\toprule
\textbf{Dataset}               & \textbf{Model} & \textbf{TPE$\downarrow$} & {\textbf{\#E$\downarrow$}} & \textbf{RAM$\downarrow$} & \textbf{VRAM$\downarrow$} \\ \midrule
\multirow{4}{*}{\textbf{ACM}}  & HAN   & \underline{0.062}   & {\underline{21.6}} & \textbf{0.855}     & \underline{2.649}      \\
                      & MAGNN & 0.151   & {45.0} & 4.967     & 5.205      \\
                      & GTN   & 24.769 & {24.4} & 8.342    & -        \\
                      & MECCH & \textbf{0.049}   & {\textbf{4.0}} & \underline{0.905}     & \textbf{1.727}      \\ \midrule
\multirow{4}{*}{\textbf{DBLP}} & HAN   & \textbf{0.068}   & {41.6} & \textbf{3.671}     & \textbf{2.937}      \\
                      & MAGNN & (7.022)   & {(22.2)} & (4.265)     & (1.569)      \\
                      & GTN   & 253.875 & {\underline{21.4}} & 55.734    & -        \\
                      & MECCH & \underline{0.181}   & {\textbf{4.2}} & \underline{4.142}     & \underline{3.645}      \\ \bottomrule
\end{tabular}
\end{table}

\begin{table*}[t]
\centering
\caption{Ablation study for node classification (IMDB, ACM, and DBLP) and link prediction (LastFM and PubMed).}
\label{tab:experiment_ablation}
\begin{tabular}{@{}ccccccccc@{}}
\toprule
\multirow{2}{*}{\textbf{Model}} & \multicolumn{2}{c}{\textbf{IMDB}}  & \multicolumn{2}{c}{\textbf{ACM}}   & \multicolumn{2}{c}{\textbf{DBLP}}   & \textbf{LastFM}   & \textbf{PubMed} \\ \cmidrule(l){2-9}
                      & Macro-F1    & Micro-F1    & Macro-F1    & Micro-F1    & Macro-F1    & Micro-F1    & ROC-AUC    & ROC-AUC    \\ \midrule
MECCH-KHop              & 53.53 & 56.28 & 91.32 & 91.26 & 93.74 & 94.52 & 83.29 & 72.74 \\
{MECCH-HAN}   & {60.03} & {62.25} & {91.55} & {91.48} & {92.67} & {93.58} & {83.34} & {73.42} \\
{MECCH-MAGNN} & {59.42} & {61.44} & {90.59} & {90.51} & {94.05} & {94.71} & {80.97} & {70.92} \\ \hdashline
MECCH-ACE               & \textbf{62.70} & 64.22 & 90.74 & 90.68 & 92.42 & 93.30 & 82.44 & 75.25 \\ \hdashline
MECCH-MMF               & 61.90 & 64.12 & 92.36 & \underline{92.32} & 94.14 & \underline{94.89} & \underline{84.22} & 75.76 \\
{MECCH-nAMF}   & {61.58} & {63.23} & {91.86} & {91.76} & {94.01} & {94.67} & {83.10} & {75.38} \\
{MECCH-gAMF}   & {62.17} & {\underline{64.25}} & {\underline{92.38}} & {92.30} & {\textbf{94.39}} & {94.86} & {84.15} & {\underline{76.26}} \\ \midrule
MECCH                   & \underline{62.59} & \textbf{64.62} & \textbf{92.74} & \textbf{92.67} & \underline{94.34} & \textbf{95.08} & \textbf{84.27} & \textbf{76.69} \\ \bottomrule
\end{tabular}
\end{table*}

\subsection{Time and Memory Usage (RQ3)}

Besides comparing HGNNs based on prediction accuracy, we also evaluate models in terms of computational efficiency.
Table~\ref{tab:efficiency} records the time spent per epoch{, the number of epochs used (excluding the patience epochs),} and CPU/GPU memory consumption of MECCH and other metapath-based HGNNs. 
All models except MAGNN on DBLP run in full-batch.

As shown, \textbf{MECCH is much more efficient than MAGNN and GTN} in training time and memory usage, with a better or comparable resource usage compared to HAN.
This is accomplished when MECCH incorporates more information than HAN (i.e., the intermediate nodes along metapaths).
MAGNN on DBLP can only run in mini-batches due to its high GPU memory usage. Hence its reported values cannot be compared directly.
On both ACM and DBLP, GTN cannot fit in the limited GPU memory. Instead, it runs on CPU with a significantly longer running time.
These results show that MECCH can achieve both superior model performance and decent computational efficiency, further proving the advantages of leveraging metapath contexts over metapath-guided neighbors and metapath instances.

\begin{figure}[t]
\centering
    \begin{subfigure}[b]{0.49\columnwidth}
        \centering
        \includegraphics[width=\linewidth]{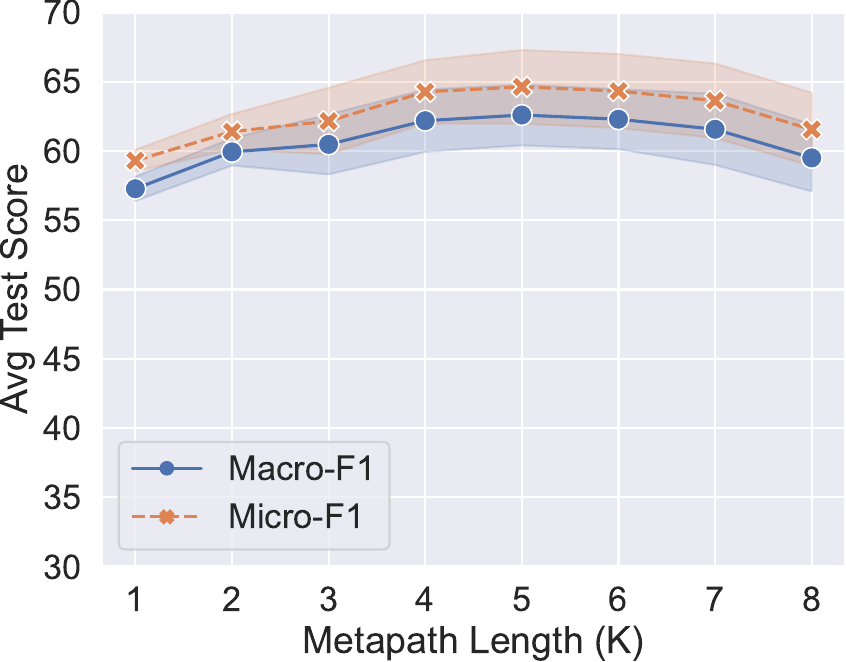}
        \caption{The metapath length ($K$).}
        \label{fig:hyperparam_K}
    \end{subfigure}
    \hfill
    \begin{subfigure}[b]{0.49\columnwidth}
        \centering
        \includegraphics[width=\linewidth]{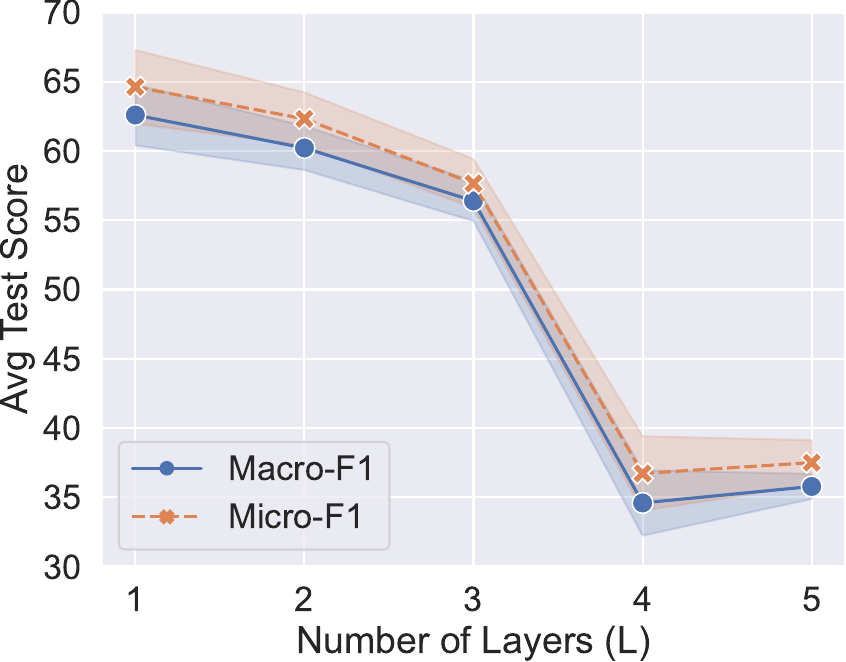}
        \caption{The number of network layers ($L$).}
        \label{fig:hyperparam_L}
    \end{subfigure}
\caption{MECCH performance on the IMDB dataset with varying metapath lengths ($K$) and model depths ($L$).}
\label{fig:hyperparam}
\end{figure}

\begin{figure*}[t]
    \centering
    \begin{subfigure}[b]{0.2\textwidth}
        \centering
        \includegraphics[width=\textwidth]{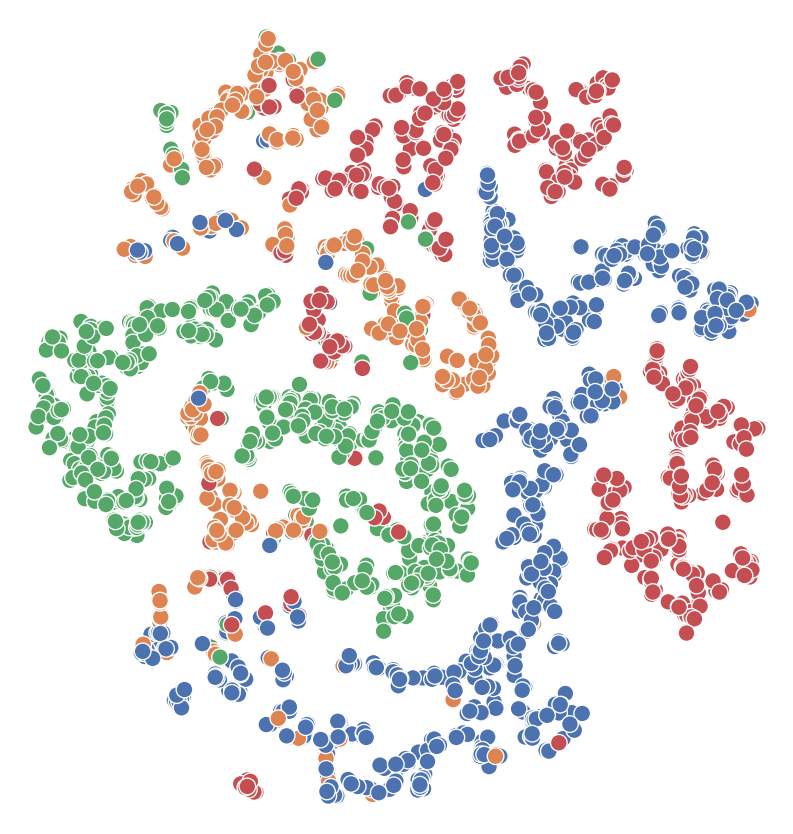}
        \caption{RGCN}
    \end{subfigure}
    \hfill
    \begin{subfigure}[b]{0.2\textwidth}
        \centering
        \includegraphics[width=\textwidth]{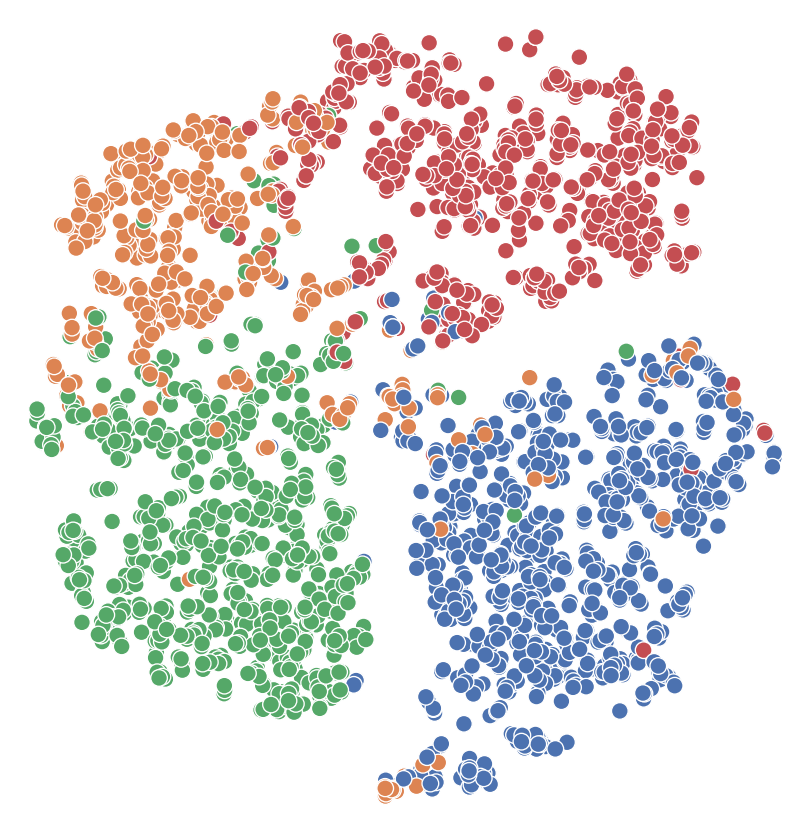}
        \caption{HGT}
    \end{subfigure}
    \hfill
    \begin{subfigure}[b]{0.2\textwidth}
        \centering
        \includegraphics[width=\textwidth]{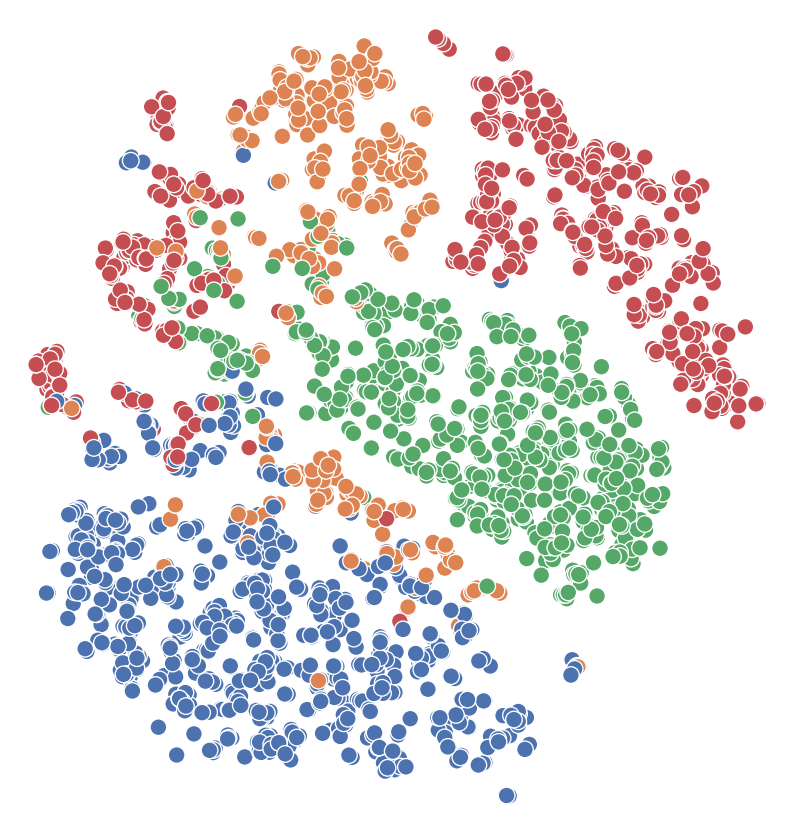}
        \caption{HAN}
    \end{subfigure}
    \hfill
    \begin{subfigure}[b]{0.2\textwidth}
        \centering
        \includegraphics[width=\textwidth]{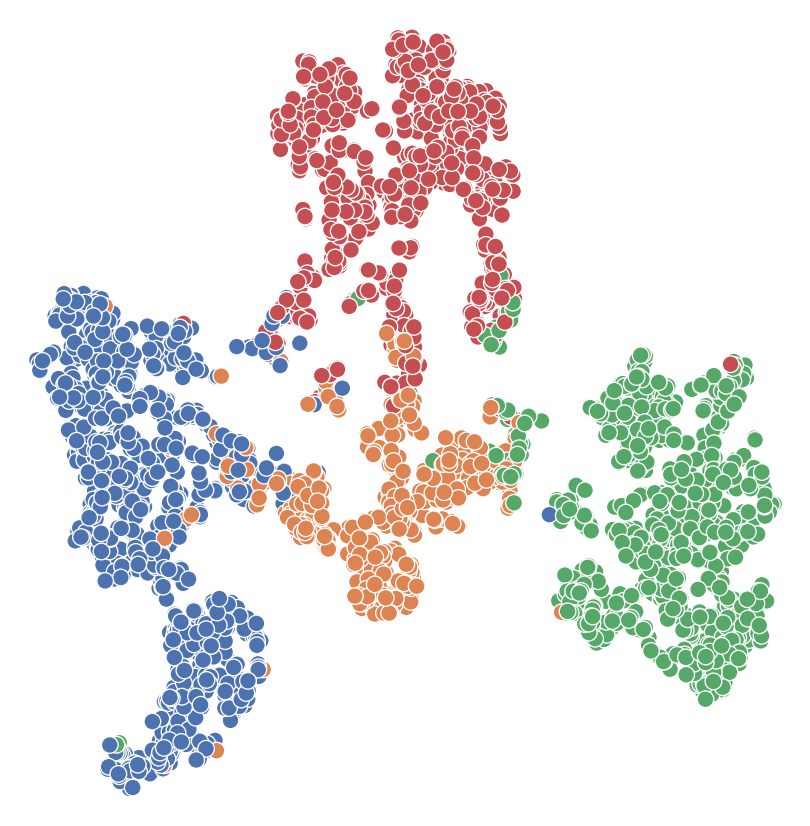}
        \caption{MECCH}
    \end{subfigure}
    \caption{Embedding visualization on DBLP. Each point is an author where color indicates the class label.}
    \label{fig:visualization_nc}
\end{figure*}

\begin{figure*}[t]
    \centering
    \begin{subfigure}[b]{0.23\textwidth}
        \centering
        \includegraphics[width=\textwidth]{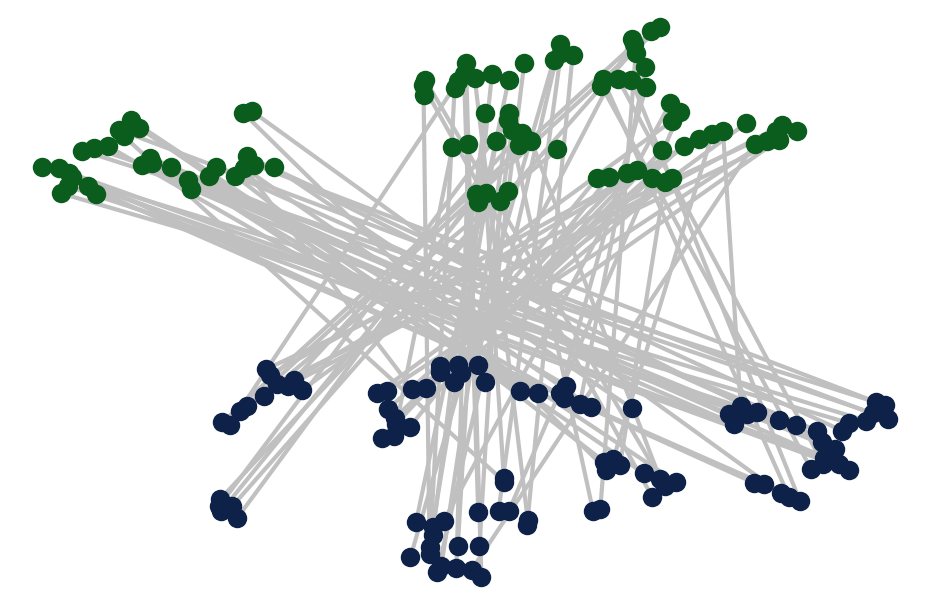}
        \caption{RGCN}
    \end{subfigure}
    \hfill
    \begin{subfigure}[b]{0.23\textwidth}
        \centering
        \includegraphics[width=\textwidth]{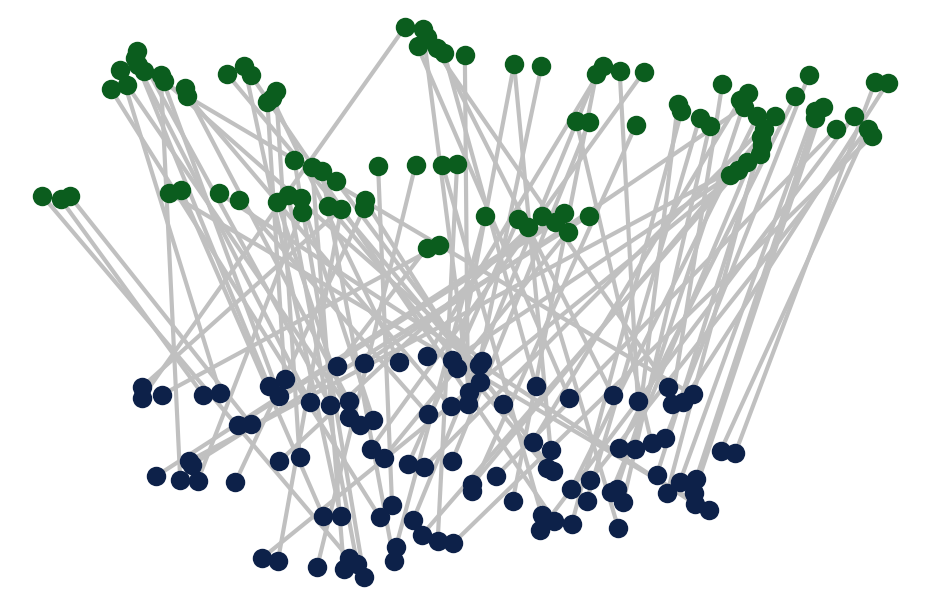}
        \caption{HGT}
    \end{subfigure}
    \hfill
    \begin{subfigure}[b]{0.23\textwidth}
        \centering
        \includegraphics[width=\textwidth]{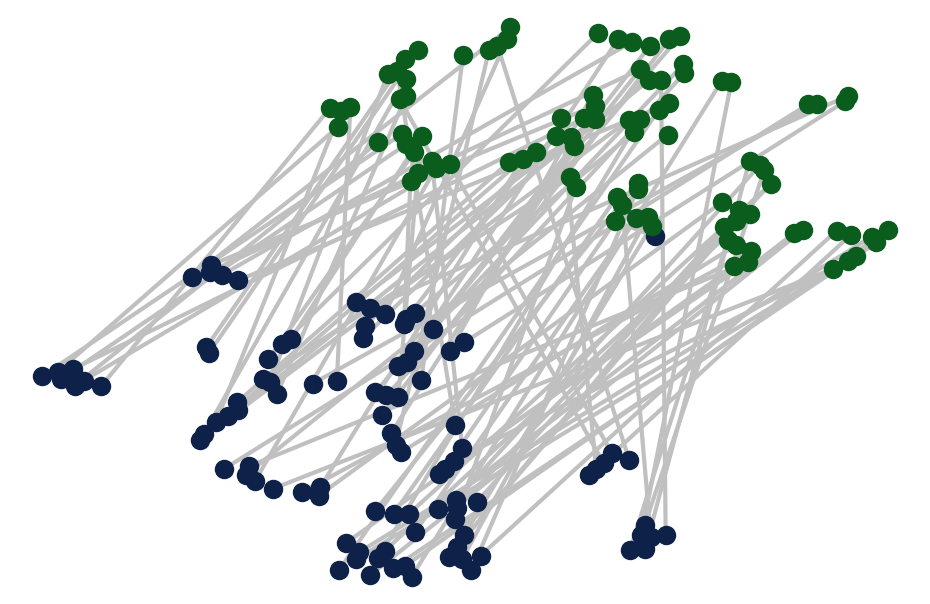}
        \caption{HAN}
    \end{subfigure}
    \hfill
    \begin{subfigure}[b]{0.23\textwidth}
        \centering
        \includegraphics[width=\textwidth]{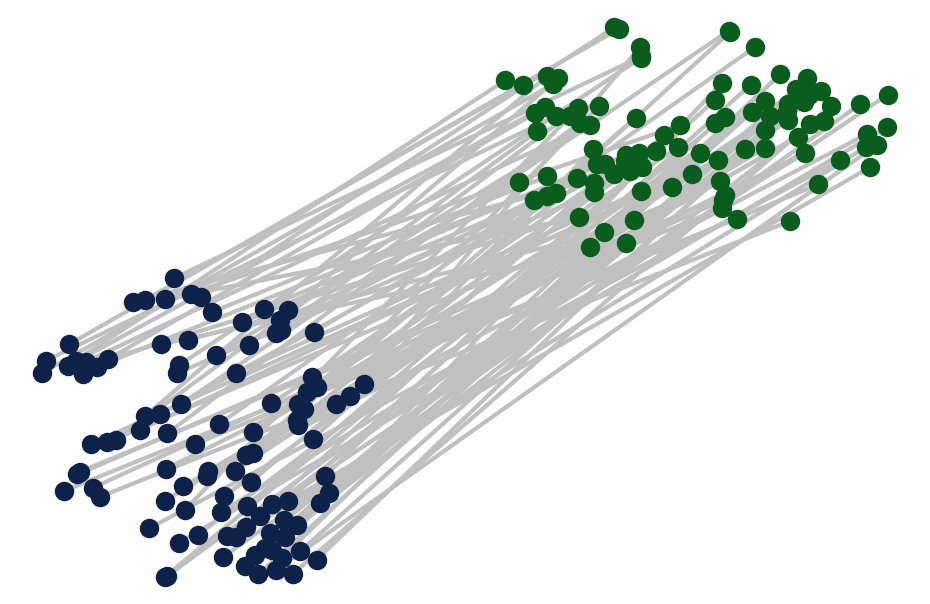}
        \caption{MECCH}
    \end{subfigure}
    \caption{Embedding visualization on LastFM. Blue and green points are users and artists, respectively.}
    \label{fig:visualization_lp}
\end{figure*}

\subsection{Ablation Study (RQ4)}

To analyze the effectiveness of the three proposed components, we conduct an ablation study by comparing our MECCH with the following model variants:
(1) \textbf{MECCH-KHop} is a variant without leveraging metapath contexts, equivalent to aggregating all the nodes within a radius of $K$ regardless of node or edge types;
{(2) \textbf{MECCH-HAN} is a variant replacing metapath contexts with metapath-guided neighbors used in HAN;}
{(3) \textbf{MECCH-MAGNN} is a variant replacing metapath contexts with metapath instances used in MAGNN;}
(4) \textbf{MECCH-ACE} is a variant using an attention-based module as the metapath context encoder;
(5) \textbf{MECCH-MMF} is a variant using the element-wise mean as the metapath fusion strategy;
{(6) \textbf{MECCH-nAMF} is a variant that fuses metapaths by applying the attention mechanism per node;}
{(7) \textbf{MECCH-gAMF} is a variant that fuses metapaths by using the attention scores computed globally based on the summaries of all nodes' representation.}
The experiment results are reported in Table~\ref{tab:experiment_ablation}, from which we have the following observations.
\begin{enumerate}
    \item By comparing MECCH against MECCH-KHop, {MECCH-HAN, and MECCH-MAGNN,} we find that MECCH can achieve a considerable performance improvement, especially on the IMDB dataset. This help to validate the superiority of leveraging metapath contexts, which is the core idea of MECCH.
    \item By comparing MECCH against MECCH-ACE, it is interesting that a simple mean graph pooling strategy is more effective for encoding metapath contexts than an attention-based module in most cases. Besides model performance, mean graph pooling as the metapath context encoder is much more efficient than other options.
    \item The difference between the results of MECCH and MECCH-MMF suggests that fusing metapaths adaptively with a 1-D convolution kernel is generally a better choice than fusing with the element-wise mean. This tells us that metapaths have different importance levels concerning the downstream task, and it is beneficial to model them.
    \item {The results of MECCH-nAMF and MECCH-gAMF suggest that the attention mechanism is not suitable for metapath fusion (MF). The node-wise attention MF in MECCH-nAMF results in different metapath weights among nodes, which is counter-intuitive and potentially jeopardises the performance. Despite having decent performance, the global attention MF in MECCH-gAMF constrains the model to only train and make inferences in full batch, because all nodes' representations are required to calculate the summary vector of each metapath. Thus, 1-D convolution is a better choice for metapath fusion, considering its simplicity, effectiveness, and efficiency.}
    \item {Using the same neighborhood subgraph definition, MECCH-MAGNN in Table~\ref{tab:experiment_ablation} outperforms or matches the original MAGNN model in Table~\ref{tab:experiment_node} and Table~\ref{tab:experiment_link}. This also indicates that the attention-free design of MECCH components improves the model performance.}
\end{enumerate}


\subsection{Hyperparameter Study}

To investigate how MECCH is affected by the hyperparameters, we conduct pilot tests on the IMDB dataset with different choices for the metapath length $K$ and the number of MECCH layers $L$.
Results are shown in Figure~\ref{fig:hyperparam}, where Figure~\ref{fig:hyperparam}~(a) shows results when $L$ is fixed at 1, and Figure~\ref{fig:hyperparam}~(b) shows results under $K=5$.
Except for the concerning hyperparameter, all other setups are kept unchanged.

From Figure~\ref{fig:hyperparam}~(a), we observe significant increase in performance when leveraging larger metapath contexts ($K > 1$).
The best results on IMDB is reached when $K$ is around 5.
Further increasing the metapath length $K$ has diminishing returns.
Obviously, the optimal choice of $K$ is closely related to the structural characteristics of the dataset and the downstream task.
Referring to Figure~\ref{fig:hyperparam}~(b), MECCH is sensitive to the choice of $L$. This makes sense because the metapath length $K$ is fixed at 5.
Adding one more MECCH layer introduces a much larger receptive field that could dramatically impact the final performance.
This also suggests that MECCH is capable of learning the complex semantics embedded in the graph with very few layers, avoiding the potential performance degradation issues when the model gets deep.

\subsection{Visualization}

To provide an intuitive assessment of the model's embedding quality, we visualize the node representations generated by different HGNNs on the DBLP and LastFM datasets.
On DBLP, we take the generated node vectors of the authors in the testing set just before applying the last GNN layer/output projection.
On LastFM, we randomly sample 100 positive user-artist pairs from the testing set and take the node vectors just before applying the DistMult decoder.
Then they are projected into a two-dimensional vector space using t-SNE.
Here we plot the t-SNE node embedding visualizations of R-GCN, HGT, HAN, and MECCH in Figure~\ref{fig:visualization_nc} for DBLP, where each color corresponds to one author class (research area), and in Figure~\ref{fig:visualization_lp} for LastFM, where blue points are users and green points are artists.

From the two figures, one can clearly observe the advantage of MECCH over other HGNNs.
Figure~\ref{fig:visualization_nc} shows that our proposed MECCH can generate more informative node representations in the node classification task.
On DBLP, MECCH makes nodes with the same class labels close to each other and nodes with different class labels far away, resulting in a clear separation between classes.
Other HGNNs would produce dispersed intra-class distribution and obscure inter-class boundaries.
Figure~\ref{fig:visualization_lp} shows that MECCH can well separate user and artist nodes and preserve the aligned correlation of the positive user-artist pairs.
While other HGNNs can roughly divide user nodes and artist nodes into two different groups, they cannot maintain good alignment between users and artists.


\section{Conclusion} \label{sec:conclusion}
In this paper, we formulate a unified framework for metapath-based HGNNs to facilitate model understanding and new model design. We analyze existing models and observe that they cannot balance model performance and computational efficiency.
To address these issues, we propose a metapath context convolution based HGNN (MECCH) with three novel components: (1) metapath context construction, (2) metapath context encoder, and (3) convolutional metapath fusion.
The three components are designed to prevent information loss and redundant computations simultaneously.

In experiments, MECCH consistently outperforms state-of-the-art baselines on five widely adopted heterogeneous graph datasets for node classification and link prediction, with improved computational efficiency.
Ablation studies validate the effectiveness of MECCH's novel components in boosting model performance.
Embedding visualizations qualitatively demonstrate the superiority of MECCH.

For future work, we plan to improve MECCH so that it can be applied to web-scale datasets, such as AMiner~\citep{DBLP:conf/kdd/TangZYLZS08} and MAG~\citep{DBLP:conf/www/SinhaSSMEHW15}, where existing metapath-based HGNNs can hardly run due to substantial computational costs.
{We also plan to incorporate various training paradigms by using MECCH as an encoder to further boost the prediction accuracy, such as self-supervised learning~\citep{DBLP:journals/corr/abs-1911-08538,DBLP:conf/aaai/ParkK0Y20}, contrastive learning~\citep{DBLP:conf/kdd/WangLHS21,DBLP:conf/kdd/Qian0WYZ22,DBLP:conf/wsdm/ChenHXWXL23}, and federated learning~\citep{DBLP:conf/ijcai/0004K23}.}

\section*{Acknowledgement}

The work described in this paper was partially supported by the National Key Research and Development Program of China, 2018AAA0100204 and the RGC General Research Fund (GRF), 2151185 (CUHK 14222922).




\bibliographystyle{elsarticle-harv} 
\bibliography{cas-refs}

\begin{thebibliography}{58}
\expandafter\ifx\csname natexlab\endcsname\relax\def\natexlab#1{#1}\fi
\providecommand{\url}[1]{\texttt{#1}}
\providecommand{\href}[2]{#2}
\providecommand{\path}[1]{#1}
\providecommand{\DOIprefix}{doi:}
\providecommand{\ArXivprefix}{arXiv:}
\providecommand{\URLprefix}{URL: }
\providecommand{\Pubmedprefix}{pmid:}
\providecommand{\doi}[1]{\href{http://dx.doi.org/#1}{\path{#1}}}
\providecommand{\Pubmed}[1]{\href{pmid:#1}{\path{#1}}}
\providecommand{\bibinfo}[2]{#2}
\ifx\xfnm\relax \def\xfnm[#1]{\unskip,\space#1}\fi
\bibitem[{Ahn et~al.(2022)Ahn, Yang, Gan, Moon and
  Wipf}]{DBLP:conf/nips/AhnYGMW22}
\bibinfo{author}{Ahn, H.}, \bibinfo{author}{Yang, Y.}, \bibinfo{author}{Gan,
  Q.}, \bibinfo{author}{Moon, T.}, \bibinfo{author}{Wipf, D.},
  \bibinfo{year}{2022}.
\newblock \bibinfo{title}{Descent steps of a relation-aware energy produce
  heterogeneous graph neural networks}, in: \bibinfo{booktitle}{NeurIPS}.
\bibitem[{Bordes et~al.(2013)Bordes, Usunier, Garc{\'{\i}}a{-}Dur{\'{a}}n,
  Weston and Yakhnenko}]{DBLP:conf/nips/BordesUGWY13}
\bibinfo{author}{Bordes, A.}, \bibinfo{author}{Usunier, N.},
  \bibinfo{author}{Garc{\'{\i}}a{-}Dur{\'{a}}n, A.}, \bibinfo{author}{Weston,
  J.}, \bibinfo{author}{Yakhnenko, O.}, \bibinfo{year}{2013}.
\newblock \bibinfo{title}{Translating embeddings for modeling multi-relational
  data}, in: \bibinfo{booktitle}{{NIPS}}, pp. \bibinfo{pages}{2787--2795}.
\bibitem[{Chang et~al.(2022)Chang, Chen, Hu, Zheng, Zhou and
  Chen}]{DBLP:journals/kbs/ChangCHZZC22}
\bibinfo{author}{Chang, Y.}, \bibinfo{author}{Chen, C.}, \bibinfo{author}{Hu,
  W.}, \bibinfo{author}{Zheng, Z.}, \bibinfo{author}{Zhou, X.},
  \bibinfo{author}{Chen, S.}, \bibinfo{year}{2022}.
\newblock \bibinfo{title}{Megnn: Meta-path extracted graph neural network for
  heterogeneous graph representation learning}.
\newblock \bibinfo{journal}{Knowl. Based Syst.} \bibinfo{volume}{235},
  \bibinfo{pages}{107611}.
\bibitem[{Chen et~al.(2018)Chen, Yin, Wang, Wang, Nguyen and
  Li}]{DBLP:conf/kdd/ChenYWWNL18}
\bibinfo{author}{Chen, H.}, \bibinfo{author}{Yin, H.}, \bibinfo{author}{Wang,
  W.}, \bibinfo{author}{Wang, H.}, \bibinfo{author}{Nguyen, Q.V.H.},
  \bibinfo{author}{Li, X.}, \bibinfo{year}{2018}.
\newblock \bibinfo{title}{{PME:} projected metric embedding on heterogeneous
  networks for link prediction}, in: \bibinfo{booktitle}{{KDD}},
  \bibinfo{publisher}{{ACM}}. pp. \bibinfo{pages}{1177--1186}.
\bibitem[{Chen et~al.(2023a)Chen, Huang, Xia, Wei, Xu and
  Luo}]{DBLP:conf/wsdm/ChenHXWXL23}
\bibinfo{author}{Chen, M.}, \bibinfo{author}{Huang, C.}, \bibinfo{author}{Xia,
  L.}, \bibinfo{author}{Wei, W.}, \bibinfo{author}{Xu, Y.},
  \bibinfo{author}{Luo, R.}, \bibinfo{year}{2023}a.
\newblock \bibinfo{title}{Heterogeneous graph contrastive learning for
  recommendation}, in: \bibinfo{booktitle}{{WSDM}}, \bibinfo{publisher}{{ACM}}.
  pp. \bibinfo{pages}{544--552}.
\bibitem[{Chen et~al.(2023b)Chen, Truong, Shen, Wang, Li, Chan and
  King}]{chen2023star}
\bibinfo{author}{Chen, Y.}, \bibinfo{author}{Truong, Q.T.},
  \bibinfo{author}{Shen, X.}, \bibinfo{author}{Wang, M.}, \bibinfo{author}{Li,
  J.}, \bibinfo{author}{Chan, J.}, \bibinfo{author}{King, I.},
  \bibinfo{year}{2023}b.
\newblock \bibinfo{title}{Topological representation learning for e-commerce
  shopping behaviors}, in: \bibinfo{booktitle}{Proceedings of the 19th
  International Workshop on Mining and Learning with Graphs (MLG)}.
\bibitem[{Chen et~al.(2023c)Chen, Zhang, Yang, Song, Ma and
  King}]{DBLP:conf/sigir/0001Z0S0K23}
\bibinfo{author}{Chen, Y.}, \bibinfo{author}{Zhang, Y.}, \bibinfo{author}{Yang,
  M.}, \bibinfo{author}{Song, Z.}, \bibinfo{author}{Ma, C.},
  \bibinfo{author}{King, I.}, \bibinfo{year}{2023}c.
\newblock \bibinfo{title}{{WSFE:} wasserstein sub-graph feature encoder for
  effective user segmentation in collaborative filtering}, in:
  \bibinfo{booktitle}{{SIGIR}}, \bibinfo{publisher}{{ACM}}. pp.
  \bibinfo{pages}{2521--2525}.
\bibitem[{Dong et~al.(2017)Dong, Chawla and Swami}]{DBLP:conf/kdd/DongCS17}
\bibinfo{author}{Dong, Y.}, \bibinfo{author}{Chawla, N.V.},
  \bibinfo{author}{Swami, A.}, \bibinfo{year}{2017}.
\newblock \bibinfo{title}{metapath2vec: Scalable representation learning for
  heterogeneous networks}, in: \bibinfo{booktitle}{{KDD}},
  \bibinfo{publisher}{{ACM}}. pp. \bibinfo{pages}{135--144}.
\bibitem[{Fu et~al.(2017)Fu, Lee and Lei}]{DBLP:conf/cikm/FuLL17}
\bibinfo{author}{Fu, T.}, \bibinfo{author}{Lee, W.}, \bibinfo{author}{Lei, Z.},
  \bibinfo{year}{2017}.
\newblock \bibinfo{title}{Hin2vec: Explore meta-paths in heterogeneous
  information networks for representation learning}, in:
  \bibinfo{booktitle}{{CIKM}}, \bibinfo{publisher}{{ACM}}. pp.
  \bibinfo{pages}{1797--1806}.
\bibitem[{Fu and King(2023)}]{DBLP:conf/ijcai/0004K23}
\bibinfo{author}{Fu, X.}, \bibinfo{author}{King, I.}, \bibinfo{year}{2023}.
\newblock \bibinfo{title}{Fedhgn: {A} federated framework for heterogeneous
  graph neural networks}, in: \bibinfo{booktitle}{{IJCAI}},
  \bibinfo{publisher}{ijcai.org}. pp. \bibinfo{pages}{3705--3713}.
\bibitem[{Fu et~al.(2020)Fu, Zhang, Meng and King}]{DBLP:conf/www/0004ZMK20}
\bibinfo{author}{Fu, X.}, \bibinfo{author}{Zhang, J.}, \bibinfo{author}{Meng,
  Z.}, \bibinfo{author}{King, I.}, \bibinfo{year}{2020}.
\newblock \bibinfo{title}{{MAGNN:} metapath aggregated graph neural network for
  heterogeneous graph embedding}, in: \bibinfo{booktitle}{{WWW}},
  \bibinfo{publisher}{{ACM} / {IW3C2}}. pp. \bibinfo{pages}{2331--2341}.
\bibitem[{Hamilton et~al.(2017)Hamilton, Ying and
  Leskovec}]{DBLP:conf/nips/HamiltonYL17}
\bibinfo{author}{Hamilton, W.L.}, \bibinfo{author}{Ying, Z.},
  \bibinfo{author}{Leskovec, J.}, \bibinfo{year}{2017}.
\newblock \bibinfo{title}{Inductive representation learning on large graphs},
  in: \bibinfo{booktitle}{{NIPS}}, pp. \bibinfo{pages}{1024--1034}.
\bibitem[{Hong et~al.(2020)Hong, Guo, Lin, Yang, Li and
  Ye}]{DBLP:conf/aaai/HongGLYLY20}
\bibinfo{author}{Hong, H.}, \bibinfo{author}{Guo, H.}, \bibinfo{author}{Lin,
  Y.}, \bibinfo{author}{Yang, X.}, \bibinfo{author}{Li, Z.},
  \bibinfo{author}{Ye, J.}, \bibinfo{year}{2020}.
\newblock \bibinfo{title}{An attention-based graph neural network for
  heterogeneous structural learning}, in: \bibinfo{booktitle}{{AAAI}},
  \bibinfo{publisher}{{AAAI} Press}. pp. \bibinfo{pages}{4132--4139}.
\bibitem[{Hu et~al.(2020)Hu, Dong, Wang and Sun}]{DBLP:conf/www/HuDWS20}
\bibinfo{author}{Hu, Z.}, \bibinfo{author}{Dong, Y.}, \bibinfo{author}{Wang,
  K.}, \bibinfo{author}{Sun, Y.}, \bibinfo{year}{2020}.
\newblock \bibinfo{title}{Heterogeneous graph transformer}, in:
  \bibinfo{booktitle}{{WWW}}, \bibinfo{publisher}{{ACM} / {IW3C2}}. pp.
  \bibinfo{pages}{2704--2710}.
\bibitem[{Kipf and Welling(2017)}]{DBLP:conf/iclr/KipfW17}
\bibinfo{author}{Kipf, T.N.}, \bibinfo{author}{Welling, M.},
  \bibinfo{year}{2017}.
\newblock \bibinfo{title}{Semi-supervised classification with graph
  convolutional networks}, in: \bibinfo{booktitle}{{ICLR} (Poster)},
  \bibinfo{publisher}{OpenReview.net}.
\bibitem[{Li et~al.(2018)Li, Han and Wu}]{DBLP:conf/aaai/LiHW18}
\bibinfo{author}{Li, Q.}, \bibinfo{author}{Han, Z.}, \bibinfo{author}{Wu, X.},
  \bibinfo{year}{2018}.
\newblock \bibinfo{title}{Deeper insights into graph convolutional networks for
  semi-supervised learning}, in: \bibinfo{booktitle}{{AAAI}},
  \bibinfo{publisher}{{AAAI} Press}. pp. \bibinfo{pages}{3538--3545}.
\bibitem[{Liang et~al.(2022)Liang, Xu, Song, King and
  Ye}]{DBLP:journals/corr/abs-2206-08181}
\bibinfo{author}{Liang, L.}, \bibinfo{author}{Xu, Z.}, \bibinfo{author}{Song,
  Z.}, \bibinfo{author}{King, I.}, \bibinfo{author}{Ye, J.},
  \bibinfo{year}{2022}.
\newblock \bibinfo{title}{Resnorm: Tackling long-tailed degree distribution
  issue in graph neural networks via normalization}.
\newblock \bibinfo{journal}{CoRR} \bibinfo{volume}{abs/2206.08181}.
\bibitem[{Lv et~al.(2021)Lv, Ding, Liu, Chen, Feng, He, Zhou, Jiang, Dong and
  Tang}]{DBLP:conf/kdd/LvDLCFHZJDT21}
\bibinfo{author}{Lv, Q.}, \bibinfo{author}{Ding, M.}, \bibinfo{author}{Liu,
  Q.}, \bibinfo{author}{Chen, Y.}, \bibinfo{author}{Feng, W.},
  \bibinfo{author}{He, S.}, \bibinfo{author}{Zhou, C.}, \bibinfo{author}{Jiang,
  J.}, \bibinfo{author}{Dong, Y.}, \bibinfo{author}{Tang, J.},
  \bibinfo{year}{2021}.
\newblock \bibinfo{title}{Are we really making much progress?: Revisiting,
  benchmarking and refining heterogeneous graph neural networks}, in:
  \bibinfo{booktitle}{{KDD}}, \bibinfo{publisher}{{ACM}}. pp.
  \bibinfo{pages}{1150--1160}.
\bibitem[{Ma et~al.(2023)Ma, Song, Hu, Li, Zhang and
  King}]{DBLP:conf/aaai/MaSHLZK23}
\bibinfo{author}{Ma, Y.}, \bibinfo{author}{Song, Z.}, \bibinfo{author}{Hu, X.},
  \bibinfo{author}{Li, J.}, \bibinfo{author}{Zhang, Y.}, \bibinfo{author}{King,
  I.}, \bibinfo{year}{2023}.
\newblock \bibinfo{title}{Graph component contrastive learning for concept
  relatedness estimation}, in: \bibinfo{booktitle}{{AAAI}},
  \bibinfo{publisher}{{AAAI} Press}. pp. \bibinfo{pages}{13362--13370}.
\bibitem[{Meng et~al.(2023a)Meng, Li, Zhao, Yu and
  King}]{DBLP:conf/sdm/MengLZ0K23}
\bibinfo{author}{Meng, Z.}, \bibinfo{author}{Li, Y.}, \bibinfo{author}{Zhao,
  P.}, \bibinfo{author}{Yu, Y.}, \bibinfo{author}{King, I.},
  \bibinfo{year}{2023}a.
\newblock \bibinfo{title}{Meta-learning with motif-based task augmentation for
  few-shot molecular property prediction}, in: \bibinfo{booktitle}{{SDM}},
  \bibinfo{publisher}{{SIAM}}. pp. \bibinfo{pages}{811--819}.
\bibitem[{Meng et~al.(2023b)Meng, Zhao, Yu and
  King}]{DBLP:conf/ijcai/MengZ0K23}
\bibinfo{author}{Meng, Z.}, \bibinfo{author}{Zhao, P.}, \bibinfo{author}{Yu,
  Y.}, \bibinfo{author}{King, I.}, \bibinfo{year}{2023}b.
\newblock \bibinfo{title}{Doubly stochastic graph-based non-autoregressive
  reaction prediction}, in: \bibinfo{booktitle}{{IJCAI}},
  \bibinfo{publisher}{ijcai.org}. pp. \bibinfo{pages}{4064--4072}.
\bibitem[{Meng et~al.(2023c)Meng, Zhao, Yu and
  King}]{DBLP:conf/ijcai/MengZ0K23a}
\bibinfo{author}{Meng, Z.}, \bibinfo{author}{Zhao, P.}, \bibinfo{author}{Yu,
  Y.}, \bibinfo{author}{King, I.}, \bibinfo{year}{2023}c.
\newblock \bibinfo{title}{A unified view of deep learning for reaction and
  retrosynthesis prediction: Current status and future challenges}, in:
  \bibinfo{booktitle}{{IJCAI}}, \bibinfo{publisher}{ijcai.org}. pp.
  \bibinfo{pages}{6723--6731}.
\bibitem[{Mikolov et~al.(2013a)Mikolov, Chen, Corrado and
  Dean}]{DBLP:journals/corr/abs-1301-3781}
\bibinfo{author}{Mikolov, T.}, \bibinfo{author}{Chen, K.},
  \bibinfo{author}{Corrado, G.}, \bibinfo{author}{Dean, J.},
  \bibinfo{year}{2013}a.
\newblock \bibinfo{title}{Efficient estimation of word representations in
  vector space}, in: \bibinfo{booktitle}{{ICLR} (Workshop Poster)}.
\bibitem[{Mikolov et~al.(2013b)Mikolov, Sutskever, Chen, Corrado and
  Dean}]{DBLP:conf/nips/MikolovSCCD13}
\bibinfo{author}{Mikolov, T.}, \bibinfo{author}{Sutskever, I.},
  \bibinfo{author}{Chen, K.}, \bibinfo{author}{Corrado, G.S.},
  \bibinfo{author}{Dean, J.}, \bibinfo{year}{2013}b.
\newblock \bibinfo{title}{Distributed representations of words and phrases and
  their compositionality}, in: \bibinfo{booktitle}{{NIPS}}, pp.
  \bibinfo{pages}{3111--3119}.
\bibitem[{Park et~al.(2020)Park, Kim, Han and Yu}]{DBLP:conf/aaai/ParkK0Y20}
\bibinfo{author}{Park, C.}, \bibinfo{author}{Kim, D.}, \bibinfo{author}{Han,
  J.}, \bibinfo{author}{Yu, H.}, \bibinfo{year}{2020}.
\newblock \bibinfo{title}{Unsupervised attributed multiplex network embedding},
  in: \bibinfo{booktitle}{{AAAI}}, \bibinfo{publisher}{{AAAI} Press}. pp.
  \bibinfo{pages}{5371--5378}.
\bibitem[{Perozzi et~al.(2014)Perozzi, Al{-}Rfou and
  Skiena}]{DBLP:conf/kdd/PerozziAS14}
\bibinfo{author}{Perozzi, B.}, \bibinfo{author}{Al{-}Rfou, R.},
  \bibinfo{author}{Skiena, S.}, \bibinfo{year}{2014}.
\newblock \bibinfo{title}{Deepwalk: online learning of social representations},
  in: \bibinfo{booktitle}{{KDD}}, \bibinfo{publisher}{{ACM}}. pp.
  \bibinfo{pages}{701--710}.
\bibitem[{Qian et~al.(2022)Qian, Zhang, Wen, Ye and
  Zhang}]{DBLP:conf/kdd/Qian0WYZ22}
\bibinfo{author}{Qian, Y.}, \bibinfo{author}{Zhang, Y.}, \bibinfo{author}{Wen,
  Q.}, \bibinfo{author}{Ye, Y.}, \bibinfo{author}{Zhang, C.},
  \bibinfo{year}{2022}.
\newblock \bibinfo{title}{Rep2vec: Repository embedding via heterogeneous graph
  adversarial contrastive learning}, in: \bibinfo{booktitle}{{KDD}},
  \bibinfo{publisher}{{ACM}}. pp. \bibinfo{pages}{1390--1400}.
\bibitem[{Ren et~al.(2019)Ren, Liu, Huang, Dai, Bo and
  Zhang}]{DBLP:journals/corr/abs-1911-08538}
\bibinfo{author}{Ren, Y.}, \bibinfo{author}{Liu, B.}, \bibinfo{author}{Huang,
  C.}, \bibinfo{author}{Dai, P.}, \bibinfo{author}{Bo, L.},
  \bibinfo{author}{Zhang, J.}, \bibinfo{year}{2019}.
\newblock \bibinfo{title}{Heterogeneous deep graph infomax}.
\newblock \bibinfo{journal}{CoRR} \bibinfo{volume}{abs/1911.08538}.
\bibitem[{Schlichtkrull et~al.(2018)Schlichtkrull, Kipf, Bloem, van~den Berg,
  Titov and Welling}]{DBLP:conf/esws/SchlichtkrullKB18}
\bibinfo{author}{Schlichtkrull, M.S.}, \bibinfo{author}{Kipf, T.N.},
  \bibinfo{author}{Bloem, P.}, \bibinfo{author}{van~den Berg, R.},
  \bibinfo{author}{Titov, I.}, \bibinfo{author}{Welling, M.},
  \bibinfo{year}{2018}.
\newblock \bibinfo{title}{Modeling relational data with graph convolutional
  networks}, in: \bibinfo{booktitle}{{ESWC}}, \bibinfo{publisher}{Springer}.
  pp. \bibinfo{pages}{593--607}.
\bibitem[{Shi et~al.(2019)Shi, Hu, Zhao and Yu}]{DBLP:journals/tkde/ShiHZY19}
\bibinfo{author}{Shi, C.}, \bibinfo{author}{Hu, B.}, \bibinfo{author}{Zhao,
  W.X.}, \bibinfo{author}{Yu, P.S.}, \bibinfo{year}{2019}.
\newblock \bibinfo{title}{Heterogeneous information network embedding for
  recommendation}.
\newblock \bibinfo{journal}{{IEEE} Trans. Knowl. Data Eng.}
  \bibinfo{volume}{31}, \bibinfo{pages}{357--370}.
\bibitem[{Sinha et~al.(2015)Sinha, Shen, Song, Ma, Eide, Hsu and
  Wang}]{DBLP:conf/www/SinhaSSMEHW15}
\bibinfo{author}{Sinha, A.}, \bibinfo{author}{Shen, Z.}, \bibinfo{author}{Song,
  Y.}, \bibinfo{author}{Ma, H.}, \bibinfo{author}{Eide, D.},
  \bibinfo{author}{Hsu, B.P.}, \bibinfo{author}{Wang, K.},
  \bibinfo{year}{2015}.
\newblock \bibinfo{title}{An overview of microsoft academic service {(MAS)} and
  applications}, in: \bibinfo{booktitle}{{WWW} (Companion Volume)},
  \bibinfo{publisher}{{ACM}}. pp. \bibinfo{pages}{243--246}.
\bibitem[{Song and King(2022)}]{DBLP:conf/aaai/SongK22}
\bibinfo{author}{Song, Z.}, \bibinfo{author}{King, I.}, \bibinfo{year}{2022}.
\newblock \bibinfo{title}{Hierarchical heterogeneous graph attention network
  for syntax-aware summarization}, in: \bibinfo{booktitle}{{AAAI}},
  \bibinfo{publisher}{{AAAI} Press}. pp. \bibinfo{pages}{11340--11348}.
\bibitem[{Sun et~al.(2019)Sun, Deng, Nie and Tang}]{DBLP:conf/iclr/SunDNT19}
\bibinfo{author}{Sun, Z.}, \bibinfo{author}{Deng, Z.}, \bibinfo{author}{Nie,
  J.}, \bibinfo{author}{Tang, J.}, \bibinfo{year}{2019}.
\newblock \bibinfo{title}{Rotate: Knowledge graph embedding by relational
  rotation in complex space}, in: \bibinfo{booktitle}{{ICLR} (Poster)},
  \bibinfo{publisher}{OpenReview.net}.
\bibitem[{Tang et~al.(2008)Tang, Zhang, Yao, Li, Zhang and
  Su}]{DBLP:conf/kdd/TangZYLZS08}
\bibinfo{author}{Tang, J.}, \bibinfo{author}{Zhang, J.}, \bibinfo{author}{Yao,
  L.}, \bibinfo{author}{Li, J.}, \bibinfo{author}{Zhang, L.},
  \bibinfo{author}{Su, Z.}, \bibinfo{year}{2008}.
\newblock \bibinfo{title}{Arnetminer: extraction and mining of academic social
  networks}, in: \bibinfo{booktitle}{{KDD}}, \bibinfo{publisher}{{ACM}}. pp.
  \bibinfo{pages}{990--998}.
\bibitem[{Vaswani et~al.(2017)Vaswani, Shazeer, Parmar, Uszkoreit, Jones,
  Gomez, Kaiser and Polosukhin}]{DBLP:conf/nips/VaswaniSPUJGKP17}
\bibinfo{author}{Vaswani, A.}, \bibinfo{author}{Shazeer, N.},
  \bibinfo{author}{Parmar, N.}, \bibinfo{author}{Uszkoreit, J.},
  \bibinfo{author}{Jones, L.}, \bibinfo{author}{Gomez, A.N.},
  \bibinfo{author}{Kaiser, L.}, \bibinfo{author}{Polosukhin, I.},
  \bibinfo{year}{2017}.
\newblock \bibinfo{title}{Attention is all you need}, in:
  \bibinfo{booktitle}{{NIPS}}, pp. \bibinfo{pages}{5998--6008}.
\bibitem[{Velickovic et~al.(2018)Velickovic, Cucurull, Casanova, Romero,
  Li{\`{o}} and Bengio}]{DBLP:conf/iclr/VelickovicCCRLB18}
\bibinfo{author}{Velickovic, P.}, \bibinfo{author}{Cucurull, G.},
  \bibinfo{author}{Casanova, A.}, \bibinfo{author}{Romero, A.},
  \bibinfo{author}{Li{\`{o}}, P.}, \bibinfo{author}{Bengio, Y.},
  \bibinfo{year}{2018}.
\newblock \bibinfo{title}{Graph attention networks}, in:
  \bibinfo{booktitle}{{ICLR} (Poster)}, \bibinfo{publisher}{OpenReview.net}.
\bibitem[{Wang et~al.(2022)Wang, Zhou, Yu, Chen, Li, Feng and
  Chen}]{DBLP:conf/www/00010YCLF022}
\bibinfo{author}{Wang, C.}, \bibinfo{author}{Zhou, S.}, \bibinfo{author}{Yu,
  K.}, \bibinfo{author}{Chen, D.}, \bibinfo{author}{Li, B.},
  \bibinfo{author}{Feng, Y.}, \bibinfo{author}{Chen, C.}, \bibinfo{year}{2022}.
\newblock \bibinfo{title}{Collaborative knowledge distillation for
  heterogeneous information network embedding}, in: \bibinfo{booktitle}{{WWW}},
  \bibinfo{publisher}{{ACM}}. pp. \bibinfo{pages}{1631--1639}.
\bibitem[{Wang et~al.(2021a)Wang, Gao, Huang, Liu, Ma and
  Vosoughi}]{DBLP:conf/aaai/WangGHLMV21}
\bibinfo{author}{Wang, L.}, \bibinfo{author}{Gao, C.}, \bibinfo{author}{Huang,
  C.}, \bibinfo{author}{Liu, R.}, \bibinfo{author}{Ma, W.},
  \bibinfo{author}{Vosoughi, S.}, \bibinfo{year}{2021}a.
\newblock \bibinfo{title}{Embedding heterogeneous networks into hyperbolic
  space without meta-path}, in: \bibinfo{booktitle}{{AAAI}},
  \bibinfo{publisher}{{AAAI} Press}. pp. \bibinfo{pages}{10147--10155}.
\bibitem[{Wang et~al.(2019a)Wang, Ji, Shi, Wang, Ye, Cui and
  Yu}]{DBLP:conf/www/WangJSWYCY19}
\bibinfo{author}{Wang, X.}, \bibinfo{author}{Ji, H.}, \bibinfo{author}{Shi,
  C.}, \bibinfo{author}{Wang, B.}, \bibinfo{author}{Ye, Y.},
  \bibinfo{author}{Cui, P.}, \bibinfo{author}{Yu, P.S.}, \bibinfo{year}{2019}a.
\newblock \bibinfo{title}{Heterogeneous graph attention network}, in:
  \bibinfo{booktitle}{{WWW}}, \bibinfo{publisher}{{ACM}}. pp.
  \bibinfo{pages}{2022--2032}.
\bibitem[{Wang et~al.(2021b)Wang, Liu, Han and Shi}]{DBLP:conf/kdd/WangLHS21}
\bibinfo{author}{Wang, X.}, \bibinfo{author}{Liu, N.}, \bibinfo{author}{Han,
  H.}, \bibinfo{author}{Shi, C.}, \bibinfo{year}{2021}b.
\newblock \bibinfo{title}{Self-supervised heterogeneous graph neural network
  with co-contrastive learning}, in: \bibinfo{booktitle}{{KDD}},
  \bibinfo{publisher}{{ACM}}. pp. \bibinfo{pages}{1726--1736}.
\bibitem[{Wang et~al.(2019b)Wang, Zhang and Shi}]{DBLP:conf/aaai/WangZS19a}
\bibinfo{author}{Wang, X.}, \bibinfo{author}{Zhang, Y.}, \bibinfo{author}{Shi,
  C.}, \bibinfo{year}{2019}b.
\newblock \bibinfo{title}{Hyperbolic heterogeneous information network
  embedding}, in: \bibinfo{booktitle}{{AAAI}}, \bibinfo{publisher}{{AAAI}
  Press}. pp. \bibinfo{pages}{5337--5344}.
\bibitem[{Yang et~al.(2015)Yang, Yih, He, Gao and
  Deng}]{DBLP:journals/corr/YangYHGD14a}
\bibinfo{author}{Yang, B.}, \bibinfo{author}{Yih, W.}, \bibinfo{author}{He,
  X.}, \bibinfo{author}{Gao, J.}, \bibinfo{author}{Deng, L.},
  \bibinfo{year}{2015}.
\newblock \bibinfo{title}{Embedding entities and relations for learning and
  inference in knowledge bases}, in: \bibinfo{booktitle}{{ICLR} (Poster)}.
\bibitem[{Yang et~al.(2022)Yang, Xiao, Zhang, Sun and
  Han}]{DBLP:journals/tkde/YangXZSH22}
\bibinfo{author}{Yang, C.}, \bibinfo{author}{Xiao, Y.}, \bibinfo{author}{Zhang,
  Y.}, \bibinfo{author}{Sun, Y.}, \bibinfo{author}{Han, J.},
  \bibinfo{year}{2022}.
\newblock \bibinfo{title}{Heterogeneous network representation learning: {A}
  unified framework with survey and benchmark}.
\newblock \bibinfo{journal}{{IEEE} Trans. Knowl. Data Eng.}
  \bibinfo{volume}{34}, \bibinfo{pages}{4854--4873}.
\bibitem[{Yang et~al.(2023a)Yang, Zhou, Pan and King}]{DBLP:conf/kdd/00010PK23}
\bibinfo{author}{Yang, M.}, \bibinfo{author}{Zhou, M.}, \bibinfo{author}{Pan,
  L.}, \bibinfo{author}{King, I.}, \bibinfo{year}{2023}a.
\newblock \bibinfo{title}{{\(\kappa\)}hgcn: Tree-likeness modeling via
  continuous and discrete curvature learning}, in: \bibinfo{booktitle}{{KDD}},
  \bibinfo{publisher}{{ACM}}. pp. \bibinfo{pages}{2965--2977}.
\bibitem[{Yang et~al.(2023b)Yang, Zhou, Xiong and
  King}]{DBLP:journals/tkde/YangZXK23}
\bibinfo{author}{Yang, M.}, \bibinfo{author}{Zhou, M.}, \bibinfo{author}{Xiong,
  H.}, \bibinfo{author}{King, I.}, \bibinfo{year}{2023}b.
\newblock \bibinfo{title}{Hyperbolic temporal network embedding}.
\newblock \bibinfo{journal}{{IEEE} Trans. Knowl. Data Eng.}
  \bibinfo{volume}{35}, \bibinfo{pages}{11489--11502}.
\bibitem[{Yang et~al.(2023c)Yang, Zhou, Ying, Chen and
  King}]{DBLP:conf/icml/00010Y0K23}
\bibinfo{author}{Yang, M.}, \bibinfo{author}{Zhou, M.}, \bibinfo{author}{Ying,
  R.}, \bibinfo{author}{Chen, Y.}, \bibinfo{author}{King, I.},
  \bibinfo{year}{2023}c.
\newblock \bibinfo{title}{Hyperbolic representation learning: Revisiting and
  advancing}, in: \bibinfo{booktitle}{{ICML}}, \bibinfo{publisher}{{PMLR}}. pp.
  \bibinfo{pages}{39639--39659}.
\bibitem[{Yang et~al.(2023d)Yang, Guan, Li, Zhao, Cui and
  Wang}]{DBLP:journals/tkde/YangGLZCW23}
\bibinfo{author}{Yang, Y.}, \bibinfo{author}{Guan, Z.}, \bibinfo{author}{Li,
  J.}, \bibinfo{author}{Zhao, W.}, \bibinfo{author}{Cui, J.},
  \bibinfo{author}{Wang, Q.}, \bibinfo{year}{2023}d.
\newblock \bibinfo{title}{Interpretable and efficient heterogeneous graph
  convolutional network}.
\newblock \bibinfo{journal}{{IEEE} Trans. Knowl. Data Eng.}
  \bibinfo{volume}{35}, \bibinfo{pages}{1637--1650}.
\bibitem[{Yang et~al.(2021)Yang, Liu, Wang, Zhou, Gan, Wei, Zhang, Huang and
  Wipf}]{DBLP:conf/icml/YangLWZGWZHW21}
\bibinfo{author}{Yang, Y.}, \bibinfo{author}{Liu, T.}, \bibinfo{author}{Wang,
  Y.}, \bibinfo{author}{Zhou, J.}, \bibinfo{author}{Gan, Q.},
  \bibinfo{author}{Wei, Z.}, \bibinfo{author}{Zhang, Z.},
  \bibinfo{author}{Huang, Z.}, \bibinfo{author}{Wipf, D.},
  \bibinfo{year}{2021}.
\newblock \bibinfo{title}{Graph neural networks inspired by classical iterative
  algorithms}, in: \bibinfo{booktitle}{{ICML}}, \bibinfo{publisher}{{PMLR}}.
  pp. \bibinfo{pages}{11773--11783}.
\bibitem[{Yoon et~al.(2022)Yoon, Palowitch, Zelle, Hu, Salakhutdinov and
  Perozzi}]{DBLP:conf/nips/YoonPZHSP22}
\bibinfo{author}{Yoon, M.}, \bibinfo{author}{Palowitch, J.},
  \bibinfo{author}{Zelle, D.}, \bibinfo{author}{Hu, Z.},
  \bibinfo{author}{Salakhutdinov, R.}, \bibinfo{author}{Perozzi, B.},
  \bibinfo{year}{2022}.
\newblock \bibinfo{title}{Zero-shot transfer learning within a heterogeneous
  graph via knowledge transfer networks}, in: \bibinfo{booktitle}{NeurIPS}.
\bibitem[{Yu et~al.(2020)Yu, Shen, Li and
  Lerer}]{DBLP:journals/corr/abs-2011-09679}
\bibinfo{author}{Yu, L.}, \bibinfo{author}{Shen, J.}, \bibinfo{author}{Li, J.},
  \bibinfo{author}{Lerer, A.}, \bibinfo{year}{2020}.
\newblock \bibinfo{title}{Scalable graph neural networks for heterogeneous
  graphs}.
\newblock \bibinfo{journal}{CoRR} \bibinfo{volume}{abs/2011.09679}.
\bibitem[{Yu et~al.(2023)Yu, Sun, Du, Liu, Lv and
  Xiong}]{DBLP:journals/tkde/YuSDLLX23}
\bibinfo{author}{Yu, L.}, \bibinfo{author}{Sun, L.}, \bibinfo{author}{Du, B.},
  \bibinfo{author}{Liu, C.}, \bibinfo{author}{Lv, W.}, \bibinfo{author}{Xiong,
  H.}, \bibinfo{year}{2023}.
\newblock \bibinfo{title}{Heterogeneous graph representation learning with
  relation awareness}.
\newblock \bibinfo{journal}{{IEEE} Trans. Knowl. Data Eng.}
  \bibinfo{volume}{35}, \bibinfo{pages}{5935--5947}.
\bibitem[{Yun et~al.(2019)Yun, Jeong, Kim, Kang and
  Kim}]{DBLP:conf/nips/YunJKKK19}
\bibinfo{author}{Yun, S.}, \bibinfo{author}{Jeong, M.}, \bibinfo{author}{Kim,
  R.}, \bibinfo{author}{Kang, J.}, \bibinfo{author}{Kim, H.J.},
  \bibinfo{year}{2019}.
\newblock \bibinfo{title}{Graph transformer networks}, in:
  \bibinfo{booktitle}{NeurIPS}, pp. \bibinfo{pages}{11960--11970}.
\bibitem[{Zhang et~al.(2019a)Zhang, Song, Huang, Swami and
  Chawla}]{DBLP:conf/kdd/ZhangSHSC19}
\bibinfo{author}{Zhang, C.}, \bibinfo{author}{Song, D.},
  \bibinfo{author}{Huang, C.}, \bibinfo{author}{Swami, A.},
  \bibinfo{author}{Chawla, N.V.}, \bibinfo{year}{2019}a.
\newblock \bibinfo{title}{Heterogeneous graph neural network}, in:
  \bibinfo{booktitle}{{KDD}}, \bibinfo{publisher}{{ACM}}. pp.
  \bibinfo{pages}{793--803}.
\bibitem[{Zhang et~al.(2018)Zhang, Shi, Xie, Ma, King and
  Yeung}]{DBLP:conf/uai/ZhangSXMKY18}
\bibinfo{author}{Zhang, J.}, \bibinfo{author}{Shi, X.}, \bibinfo{author}{Xie,
  J.}, \bibinfo{author}{Ma, H.}, \bibinfo{author}{King, I.},
  \bibinfo{author}{Yeung, D.}, \bibinfo{year}{2018}.
\newblock \bibinfo{title}{{Ga}{AN}: Gated attention networks for learning on
  large and spatiotemporal graphs}, in: \bibinfo{booktitle}{{UAI}},
  \bibinfo{publisher}{{AUAI} Press}. pp. \bibinfo{pages}{339--349}.
\bibitem[{Zhang et~al.(2019b)Zhang, Shi, Zhao and
  King}]{DBLP:conf/ijcai/ZhangSZK19}
\bibinfo{author}{Zhang, J.}, \bibinfo{author}{Shi, X.}, \bibinfo{author}{Zhao,
  S.}, \bibinfo{author}{King, I.}, \bibinfo{year}{2019}b.
\newblock \bibinfo{title}{{STAR-GCN:} stacked and reconstructed graph
  convolutional networks for recommender systems}, in:
  \bibinfo{booktitle}{{IJCAI}}, \bibinfo{publisher}{ijcai.org}. pp.
  \bibinfo{pages}{4264--4270}.
\bibitem[{Zhang et~al.(2022)Zhang, Wang, Zhu, Shi, Zhang and
  Zhou}]{DBLP:conf/aaai/ZhangWZSZZ22}
\bibinfo{author}{Zhang, M.}, \bibinfo{author}{Wang, X.}, \bibinfo{author}{Zhu,
  M.}, \bibinfo{author}{Shi, C.}, \bibinfo{author}{Zhang, Z.},
  \bibinfo{author}{Zhou, J.}, \bibinfo{year}{2022}.
\newblock \bibinfo{title}{Robust heterogeneous graph neural networks against
  adversarial attacks}, in: \bibinfo{booktitle}{{AAAI}},
  \bibinfo{publisher}{{AAAI} Press}. pp. \bibinfo{pages}{4363--4370}.
\bibitem[{Zhao et~al.(2020)Zhao, Wang, Shi, Liu and
  Ye}]{DBLP:conf/ijcai/ZhaoWSLY20}
\bibinfo{author}{Zhao, J.}, \bibinfo{author}{Wang, X.}, \bibinfo{author}{Shi,
  C.}, \bibinfo{author}{Liu, Z.}, \bibinfo{author}{Ye, Y.},
  \bibinfo{year}{2020}.
\newblock \bibinfo{title}{Network schema preserving heterogeneous information
  network embedding}, in: \bibinfo{booktitle}{{IJCAI}},
  \bibinfo{publisher}{ijcai.org}. pp. \bibinfo{pages}{1366--1372}.
\bibitem[{Zhou et~al.(2021)Zhou, Dong, Wang, Lee, Hooi, Xu and
  Feng}]{DBLP:conf/cikm/ZhouDWLHXF21}
\bibinfo{author}{Zhou, K.}, \bibinfo{author}{Dong, Y.}, \bibinfo{author}{Wang,
  K.}, \bibinfo{author}{Lee, W.S.}, \bibinfo{author}{Hooi, B.},
  \bibinfo{author}{Xu, H.}, \bibinfo{author}{Feng, J.}, \bibinfo{year}{2021}.
\newblock \bibinfo{title}{Understanding and resolving performance degradation
  in deep graph convolutional networks}, in: \bibinfo{booktitle}{{CIKM}},
  \bibinfo{publisher}{{ACM}}. pp. \bibinfo{pages}{2728--2737}.

\end{thebibliography}





\end{document}